\documentclass{article}

\PassOptionsToPackage{numbers}{natbib}

\usepackage[preprint]{neurips_2024}

\usepackage[utf8]{inputenc} % allow utf-8 input
\usepackage[T1]{fontenc}    % use 8-bit T1 fonts
\usepackage{hyperref}       % hyperlinks
\usepackage{url}            % simple URL typesetting
\usepackage{booktabs}       % professional-quality tables
\usepackage{amsfonts}       % blackboard math symbols
\usepackage{nicefrac}       % compact symbols for 1/2, etc.
\usepackage{microtype}      % microtypography
\usepackage{xcolor}         % colors
\usepackage{wrapfig}

\usepackage[ruled,boxed]{algorithm2e} 
\usepackage{algpseudocode}

\usepackage{inconsolata}
\usepackage{xcolor}
\usepackage{multirow}
\usepackage{booktabs}
\usepackage{makecell}
\usepackage{amssymb}
\usepackage{colortbl}
\usepackage{graphicx}
\usepackage{bbding}
\usepackage{utfsym}
\usepackage{caption}
\usepackage{subcaption}
\usepackage{pifont}
\usepackage{amsmath}
\usepackage{enumitem}
\usepackage[export]{adjustbox}% http://ctan.org/pkg/adjustbox
\usepackage{xspace}

\definecolor{ForestGreen}{RGB}{34,139,34}
\definecolor{BrickRed}{rgb}{.72,0,0}
\definecolor{LakeBlue}{RGB}{0,61,153}

\definecolor{self1}{RGB}{192,0,0}
\definecolor{self2}{RGB}{46,117,182}
\definecolor{self3}{RGB}{118,113,113}

\definecolor{ao(english)}{rgb}{0.0, 0.5, 0.0}
\definecolor{veronica-red}{RGB}{196,30,58}

\usepackage{fontawesome}
\definecolor{ForestGreen}{RGB}{34,139,34}
\definecolor{BrickRed}{rgb}{.72,0,0}
\definecolor{LakeBlue}{RGB}{0,61,153}
\newcommand{\fstar}{\textsuperscript{\fontsize{6pt}{6pt}\selectfont \faStarO}}
\newcommand{\fmoon}{\textsuperscript{\fontsize{6pt}{6pt}\selectfont \faMoonO}}

\newcommand{\ours}{\textsl{ENVISIONS}\xspace}

\title{Interactive Evolution: A Neural-Symbolic Self-Training Framework For Large Language Models}

\author{Fangzhi Xu\textsuperscript{$\diamondsuit$$\heartsuit$}\thanks{\, Work done during internship at Shanghai AI Lab.}, Qiushi Sun\fstar\textsuperscript{$\heartsuit$}, Kanzhi Cheng\fmoon, Jun Liu\textsuperscript{$\diamondsuit$}, Yu Qiao\textsuperscript{$\heartsuit$}, Zhiyong Wu\textsuperscript{$\heartsuit$}\thanks{\,Corresponding Author.}\\
        \textsuperscript{$\diamondsuit$}Xi'an Jiaotong University  \: \textsuperscript{$\heartsuit$}Shanghai Artificial Intelligence Laboratory \\
        \fstar The University of Hong Kong  \: \fmoon Nanjing Univerisity  \\
        \texttt{fangzhixu98@gmail.com, wuzhiyong@pjlab.org.cn}
        }

\begin{document}

\maketitle

% \begin{abstract}
%   The abstract paragraph should be indented \nicefrac{1}{2}~inch (3~picas) on
%   both the left- and right-hand margins. Use 10~point type, with a vertical
%   spacing (leading) of 11~points.  The word \textbf{Abstract} must be centered,
%   bold, and in point size 12. Two line spaces precede the abstract. The abstract
%   must be limited to one paragraph.
% \end{abstract}

% \section{Symbol-LLM2.0}

\begin{abstract}
One of the primary driving forces contributing to the superior performance of Large Language Models (LLMs) is the extensive availability of human-annotated natural language data, which is used for alignment fine-tuning. This inspired researchers to investigate self-training methods to mitigate the extensive reliance on human annotations. However, the current success of self-training has been primarily observed in natural language scenarios, rather than in the increasingly important neural-symbolic scenarios. To this end, we propose an environment-guided neural-symbolic self-training framework named \ours. It aims to overcome two main challenges: (1) the scarcity of symbolic data, and (2) the limited proficiency of LLMs in processing symbolic language.
Extensive evaluations conducted on three distinct domains demonstrate the effectiveness of our approach. Additionally, we have conducted a comprehensive analysis to uncover the factors contributing to \ours's success, thereby offering valuable insights for future research in this area. Code will be available at \url{https://github.com/xufangzhi/ENVISIONS}.
% We are committed to making our code, models, and generated symbolic data available after acceptance.

% The success of self-training techniques for Large Language Models (LLMs) has primarily been observed in natural language (NL) , by iteratively synthesizing NL input-output pairs ($x$-$y$).
% However, the promising neural-symbolic scenarios require generating symbolic intermediate form $a$ to produce $y$ through execution, which presents new challenges for self-training techniques under the complex setting. 
% Potential solutions have obvious drawbacks, either by distilling teacher LLMs or employing reinforced methods.
% To address the challenges of (1) the scarcity of symbolic data; and (2) the limited proficiency of LLMs in neural-symbolic scenarios, we propose an \emph{Environment-guided Self-Training} approach to bootstrap LLMs through environmental interaction. Built upon it, we propose a novel neural-symbolic self-training framework \ours without human-annotated symbolic data.
% It features the ability of data synthesis and filtering through self-exploration, self-refinement, and self-rewarding strategies. 
% Optimized by the designed RL-free contrastive loss, LLMs are guided to learn from mistakes and achieve self-improvement.
% Extensive experiments across three domains verify the effectiveness of \ours.
% In-depth analysis provides new insights into the comparison and selection between SFT and reinforced loss.

\end{abstract}
\section{Introduction}

Large Language Models (LLMs)~\citep{achiam2023gpt,team2023gemini} have undergone extensive training using massive data, enabling them to possess remarkable capabilities across diverse domains. One of the main recipes of LLMs' success is the post-pretraining effort to achieve alignment with downstream tasks~\citep{alpaca,yin2023selfaware}. The effective alignment primarily relies on \textit{the accessibility of a substantial volume of expensive human-annotated data}, employing techniques such as Supervised Fine-Tuning (SFT)~\citep{ivison2023camels} or Reinforcement Learning from Human Feedback (RLHF)~\cite{ouyang2022training}. Recently, there has been a growing interest in developing self-training methods that enable fine-tuning of LLMs without human annotations, thereby reducing cost and streamlining the training process~\cite{yuan2024self}. 

Notable progress has been made in self-training techniques for natural language (NL) scenarios~\citep{chen2024self, rosset2024direct}, where researchers focus on improving LLMs by synthesizing their own natural language input-output pairs. However, in recent years, there has been a growing emphasis on delegating tasks to external tools/environments to expand the capability boundaries of LLMs. The shift in focus necessitates the generation of a symbolic intermediate representation \textit{a} that can be executed in the environment to faithfully produce the desired output \textit{y}. This neural-symbolic framework~\cite{xu2023symbol} has achieved significant success in complex planning~\cite{liu2023llm+}, mathematical reasoning~\cite{gou2023tora}, robotic planning~\cite{hu2023chain}, and agentic tasks~\cite{zheng2023synapse,wu2024copilot}.
In contrast to the abundance of NL annotation data (\textit{x-y}), curating symbolic annotation (\textit{x-a-y}) is significantly more challenging and costly due to the scarcity and inherent complexity of symbolic language (SL). In this paper, we delve into the exploration of effective self-training methods for LLMs within complex neural-symbolic scenarios, all without human-annotated symbolic data.

\begin{figure*}[t]
\large
\centering
\includegraphics[scale=0.42]{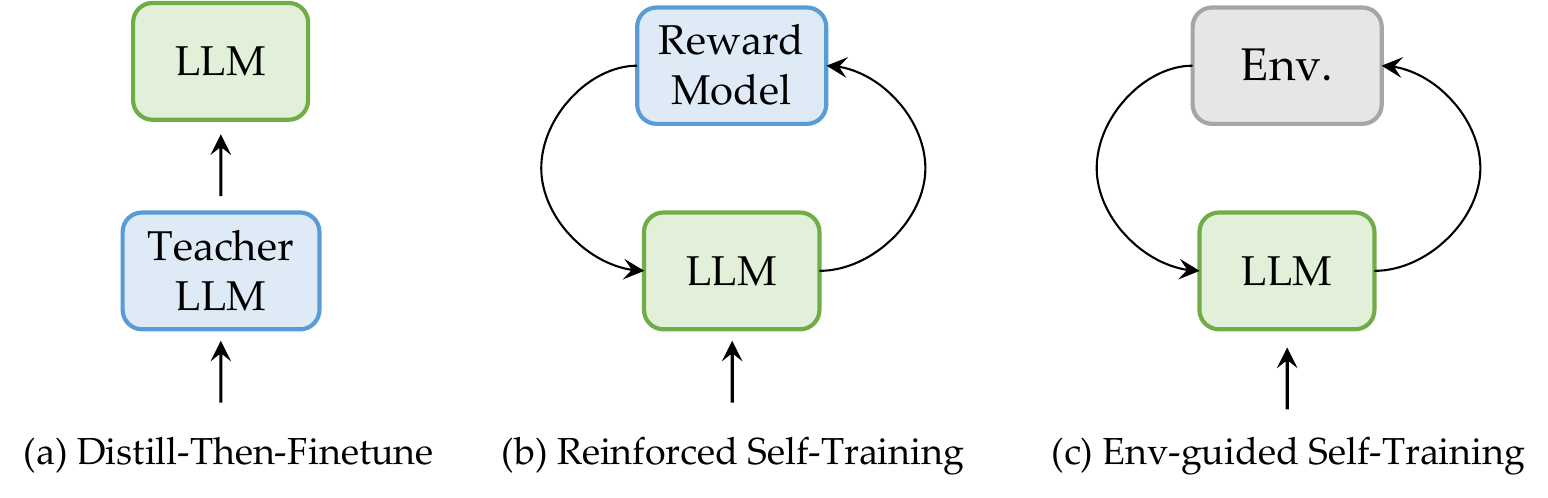}
\caption{Weak-to-strong paradigm comparison. (a) Distill-then-Finetune paradigm. (b) Reinforced Self-Training methods. (c) Env-guided Self-Training paradigm.}
\vspace{-0.4cm}
\label{paradigm}
\end{figure*}

% \vspace{-0.5cm}
% \sqs{Should we add a little bit NS background?}
% Current success in self-training falls into two categories,
% sqs: 这里他妈的绝对需要多加几个citations 
Current self-training approaches in empowering LLMs in SL-centric scenarios fall into two categories, each with its own drawbacks. \emph{Distill-then-Finetune}~\cite{ivison2023camels,xu2023wizardlm}, shown in Fig.~\ref{paradigm}(a), entails fine-tuning a less powerful LLM using distilled data obtained from a teacher LLM, such as GPT-4~\cite{achiam2023gpt}. Although this method is simple yet effective, its application is constrained by the requirement of an already existing stronger LLM and the associated high costs. Furthermore, the performance of the student LLM is upper-bounded by the capabilities of the teacher LLM. \emph{Reinforced Self-Training}~\cite{gulcehre2023reinforced,singh2023beyond}, as shown in Fig.~\ref{paradigm} (b), iteratively improves a weak LLM by leveraging reinforcement learning algorithms~\cite{rafailov2024direct}, guided by customized reward models. Nevertheless, reinforced methods are constrained by their inefficiency in training and/or reliance on human annotations for reward model training.
   
% The second paradigm is named \emph{Reinforced Self-Training}. The main idea is to iteratively optimize the weak LLM through the reinforcement learning (RL) strategy. One thread of previous works~\citep{gulcehre2023reinforced, singh2023beyond} train a reward model using human-annotated data in advance and iteratively label the generated outputs from the weak LLMs to facilitate self-training. But its reliance on human-annotated data limits the generalization to the data scarcity setting. Some other attempts~\cite{yuan2024self,chen2024self} introduce preference optimization loss (e.g., DPO~\cite{rafailov2024direct}) to make the weak LLM itself as an implicit reward model. Nevertheless, reinforced loss inherently constrains the efficiency of LLM improvement.

To address the limitations of previous approaches, 
this work focuses on two key challenges: enhancing the proficiency of LLMs in processing SL and eliminating the requirement for human-annotated data. 
Illustrated in Figure~\ref{paradigm} (c), the proposed approach, called \emph{Environment-guided (Env-guided) self-training}, involves iterative training of LLMs through interactions with an embodied environment.
Built upon the approach, 
we propose an \textbf{ENV}-gu\textbf{I}ded \textbf{S}elf-tra\textbf{I}ning framework f\textbf{O}r \textbf{N}eural \textbf{S}ymbolic scenarios, named \ours.
As an example, consider the training of LLMs for web browsing tasks, i.e., training a web agent. Given a web manipulation task $x$, the agent generates multiple candidate actions $a \in \mathcal{A}$ and executes these actions within the web browser, resulting in both correct and incorrect outcomes. A self-rewarding algorithm is designed to post-process the agent's trajectories and create contrastive training pairs. These correct-incorrect trajectory pairs, along with a self-refining loss, are utilized to empower the LLMs to self-correct and improve their performance.

Through the Env-guided self-training approach, the LLMs leverage the interactive nature of the embodied environment to generate trajectories and learn symbolic language processing abilities, mitigating the need for human annotations. Through extensive evaluation,  we found that \ours can consistently convert an existing LLM to a stronger one without reliance on existing stronger models or reward models. It's also worth noting that \ours and previous methods are not mutually exclusive, but we leave it as a future work to explore their synergy. 

We highlight our contributions as follows:

\textbf{(1) A neural-symbolic self-training framework:} 
We propose a novel framework \ours for neural-symbolic self-training. The proposed framework can eliminate the need for human annotation or a stronger teacher model during self-training. 

\textbf{(2) Comprehensive evaluations and analysis:} 
We perform extensive evaluation across three different domains to demonstrate the effectiveness of \ours over previous state-of-the-art self-training methods. Additionally, we conducted an in-depth analysis to uncover the underlying reasons for the remarkable performance of \ours and to highlight its potential as a new paradigm for neural-symbolic self-training.

\textbf{(3) Insights on Env-guided neural-symbolic self-training:} 
Our research provides valuable insights, supported by evidence, into the training process of Env-guided neural-symbolic self-training. These findings pave the way for future developments in this area of research. 
\section{Related Work}

% sqs: pls. comment my comments when issues are resolved.

\paragraph{Self-Training Methods.}
% \sqs{I suggest changing this para title to ``Self-Training w. RL''}
% \sqs{The term DPO needs to be declared in previous sections}
Self-training,
particularly when integrated with RL~\citep{schulman2017proximal,ouyang2022training}, offers a promising avenue for models to learn from their own outputs,
minimizing the need for extensive human annotations. 
% Self-training has risen as a promising approach for models to bootstrap from their outputs, 
% without the need for extensive human annotations. 
% One intuitive solution is to optimize through RL techniques (e.g., RLHF~\cite{ouyang2022training}). 
% One intuitive solution is to optimize through RL-based techniques~\citep{schulman2017proximal} (e.g., RLHF~\cite{ouyang2022training}). 
% One intuitive solution is to integrate it with RL-based techniques~\citep{schulman2017proximal} (e.g., RLHF~\cite{ouyang2022training}). 
Typically,
recent advances~\citep{gulcehre2023reinforced} leverage well-trained reward models to filter better training samples,
and further decouples the data collection and policy optimization steps for reinforced self-training~\citep{singh2023beyond,liu2023statistical}.
% ReST~\cite{gulcehre2023reinforced} 
% relies 
% leverages
% on the 
% well-trained reward models to filter better training samples. 
% \sqs{Recent advances like,}
% Recent advances like
% $\textrm{ReST}^{EM}$~\cite{singh2023beyond} further decouples the data collection and policy optimization steps for reinforced self-training. 
However, these approaches heavily rely on a 
% well-annotated % 其实就是人标数据去train的模型
strong reward model, 
which limits its applicability and training efficiency. 
Following the success of DPO~\citep{rafailov2024direct},
% Following \citet{rafailov2024direct}, 
self-play frameworks have emerged as a new path that implicitly models the preferences among unlabeled rationales in iterative DPO styles~\citep{chen2024self,rosset2024direct}. 
% \cite{chen2024self} and \cite{rosset2024direct} propose self-play frameworks
% \sqs{self-play frameworks emerge as ...} 
% in iterative DPO styles, which implicitly models the preferences among unlabeled rationales. 
More recently,
\cite{yuan2024self} incorporates the self-rewarding strategy into the DPO step, which reduces the reliance on external models. 
% But
Nevertheless,
% they are inherently limited by efficiency issues~\citep{wang2023upet}. 
such methods still face efficiency issues~\citep{wang2023upet}. 
Beyond RL,
previous works~\citep{zelikman2022star,ni2022learning} optimize policy models within 
% the correct outputs in an 
iterative SFT frameworks,
yet neglecting the value of negative samples.
% However, these approaches often ignore the value of negative samples. 
Moreover, current approaches solely integrate the neural model into self-training,
constraining the faithfulness and generalization to executable symbolic environments. 

% beyond rl, 但是想弱化这件事，还是可以和rl比，我们更加轻量化，除了star和ansong ni，其他都是rl，特点是只用正样本，而我们可以用负样本
% some previous works like ~\cite{zelikman2022star} and \cite{ni2022learning} optimize the policy model with the correct outputs in an iterative SFT framework,
% but the value of negative samples is ignored. 
% \sqs{add PPO~\citep{schulman2017proximal}}

\vspace{-0.8em}
\paragraph{Data Synthesis with LLMs.}
Obtaining high-quality human-annotated reasoning traces to optimize LLMs has been a long-standing challenge~\citep{mukherjee2023orca}.
% acquiring 
% In some scenarios,
% \sqs{what scenario? need a citation here}
% it is challenging to acquire high-quality human-annotated rationale to optimize LLMs. 
% \sqs{rationale -> reasoning traces?}
% sqs: 这里我顺序交换了一下
Beyond well-established approaches utilizing data augmentation strategies to obtain diversified training data~\citep{deng2023rephrase, lee2024llm2llm}.
Recent efforts~\citep{yue2023mammoth,zeng2023agenttuning,cheng2024seeclick} primarily distill strong LLMs~\citep{achiam2023gpt,anil2023palm} to generate novel samples in the given format. 
% Also, 
% Plenty of recent efforts~\citep{yue2023mammoth,zeng2023agenttuning,cheng2024seeclick} primarily distill strong LLMs~\citep{achiam2023gpt,anil2023palm} to generate novel samples in the given format. 
% Also, 
% some previous works~\cite{deng2023rephrase, lee2024llm2llm} apply data augmentation strategies to obtain diversified training data.
They either generate more diverse samples from seed data through self-instruct~\citep{wang2023selfinstruct} or enhance diversity through sample rephrasing and rewriting~\citep{wei2023symbol,xu2023large}.
Combining environments, learning from the trajectory of LLMs for behavior cloning has also emerged as a promising approach~\citep{lifshitz2023steve,zhao2024really}.
However, current works mainly employ close-sourced LLMs (e.g., GPT-4) for data synthesis, which is a cost. 
We propose to achieve data bootstrapping from weak LLMs without relying on external models.
% \sqs{behavior clone env-based~\citep{lifshitz2023steve}}

% sqs: 把compiler based的东西加进来也行
% \paragraph{Neural-Symbolic\sqs{Symbolism?} for LLMs}

\vspace{-0.8em}
\paragraph{Neural-Symbolic Integration for LLMs.}
% In the era of LLM, 
Neural-symbolic methods synergize the powerful generation capacity of LLMs with the reliability and interpretability 
% inherent in 
of
symbolic systems. 
% Unlike CoT~\cite{wei2022chain},
% which purely relies on NL reasoning chains,
Typically,
PAL/PoT~\cite{gao2023pal,chen2022program} synthesize executable programs as intermediate reasoning steps to solve numerical problems.
This strategy of delegating problems to external solvers (e.g., Python interpreter), 
has gained significant traction~\citep{sun2024survey}.
ToRA~\cite{gou2023tora} and Logic-LM~\cite{pan2023logic} apply neural code generation and symbolic execution on math and logical reasoning respectively, enhanced with necessary refinement.
Moreover,
Symbol-LLM~\cite{xu2023symbol} unifies neural-symbolic applications under a \emph{Symbol+Delegation} setting. 
% Symbol-LLM~\cite{xu2023symbol} optimizes plenty of symbolic generation and unifies neural-symbolic applications as \emph{Symbol+Delegation}. 
Beyond reasoning,
recent endeavors have extended the application into agent scenarios~\cite{xu2023tool,qin2023toolllm} and leverage external feedback from the environment~\cite{zheng2023synapse,yang2024react} for refinement. 
% \cite{schick2024toolformer, tang2023toolalpaca} stress the generation of function calls to facilitate the utilization of external tools. 
% \cite{zheng2023synapse,yang2024react} receive feedback through interaction with the environment to improve the neural generation steps.
% However, these approaches focus on optimizing the use of LLMs rather than offering autonomous self-improvement paradigms for them. 
However, 
these approaches mainly optimize LLM usage rather than providing autonomous self-improvement.

\section{Methodology}

\begin{figure*}[t]
\large
\centering
\includegraphics[scale=0.55]{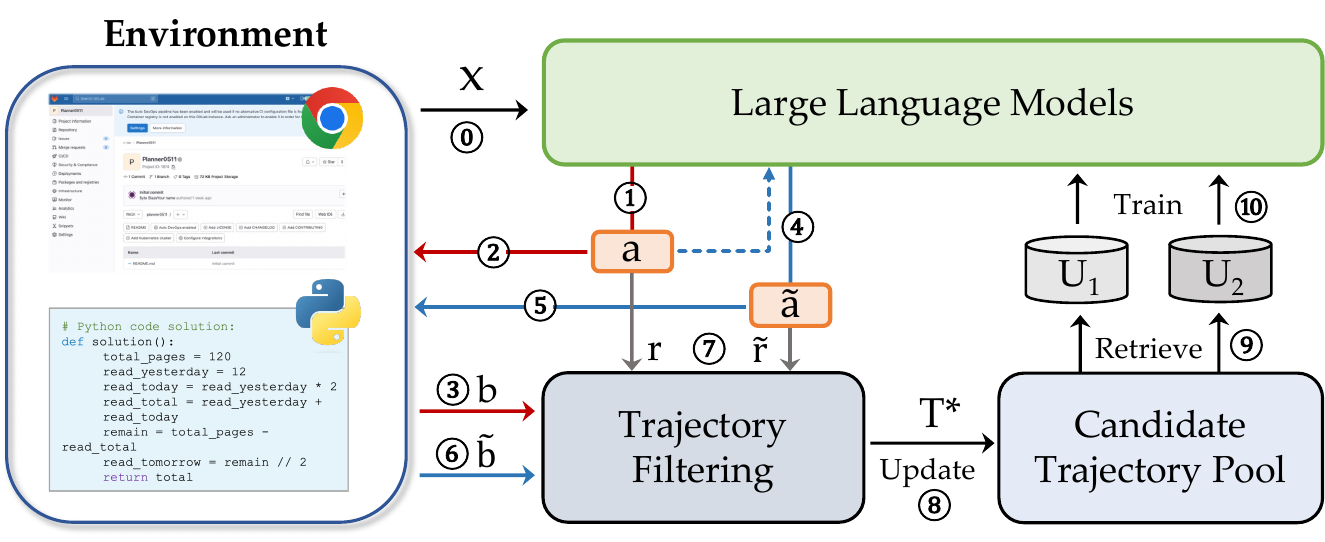}
\caption{The neural-symbolic self-training framework \ours. \textcolor{self1}{$\rightarrow$} denotes \emph{self-exploration} process (Step \ding{172}-\ding{174}), \textcolor{self2}{$\rightarrow$} indicates \emph{self-refinement} (Step \ding{175}-\ding{177}), and \textcolor{self3}{$\rightarrow$} is \emph{self-rewarding} (Step \ding{178}).}
\label{model}
\vspace{-1em}
\end{figure*}

\subsection{Preliminaries}

In neural-symbolic scenarios, based on the NL input $x$, LLMs are required to produce symbolic solution $a$ to obtain the desired output $y$ through the execution in the environment $\textbf{\textsc{Env}}$. 
To adapt the weak LLMs to such complex settings and curate extensive $(x,a,y)$ pairs, we propose to iteratively interact with $\textbf{\textsc{Env}}$, thus boosting LLMs through self-training. 
% In light of the challenges, we propose to convert LLMs from weak to strong in an iterative self-training manner. 
For each iteration $i$, the LLM $\pi_{\theta_i}$ will be provided with the task data set $\{(x^{(i)}, y^{(i)})\}$, with $J$ input-output pairs. 
Without loss of generality, we assume the samples keep static between iterations.

%sqs: em?这个开头怎么有点奇怪，我们框架的核心是构建...更强的LLMs
% The core of our framework\sqs{motivations?} is to build stronger LLMs by iteratively interacting with the environments and conducting self-training. 
% The environments (e.g., web browser in agent tasks), formulated as $\textbf{\textsc{Env}}$, can give feedback after execution on the inputs (e.g., click command)\ckz{how does the web browser give feedback? maybe use the solver/interpreter as a clearer example}. 
% Also, active memory $\textsc{M}_{act}$ and long-term memory $\textsc{M}_{l}$ are involved to supplement the self-training.\zy{confusing}
% We designate the process of interacting with the environment and memories as \emph{symbolic}\zy{process as symbolic?}. 
% For each iteration $i$, the task data set $\{(x_j^{(i)}, y_j^{(i)})\}$ \zy{what is j here?}can either be static samples or dynamically derived from $\textbf{\textsc{Env}}$. \zy{?}
% $x_j^{(i)}$ is a task instruction or query in the natural language form, while $y_j^{(i)}$ is a successful final state (agent tasks) or correct answer (reasoning tasks).
% To interact smoothly with the environment, policy model $\pi_{\theta_i}$ is required to generate the executable symbolic output $a_j^{(i)}$ via the neural process $\pi_{\theta_i}(x_j^{(i)})$. The ultimate goal is to develop the stronger policy $\pi_{\theta}$ from the iterations. \sqs{waiting for xfz}

Figure~\ref{model} presents the iterative self-training framework, named \ours. Our framework is specifically designed to address two key challenges: (1) the scarcity of SL data and (2) the limited proficiency of LLMs in SL.
Data scarcity limitation is addressed by the online exploration stage (Step \ding{172}-\ding{178}).
To convert LLMs from weak to strong in addressing SL, we employ LLM training using a carefully designed loss function and filtered data (Step \ding{179}-\ding{181}).
To simplify the expression, we omit the indicator of iteration $i$ in the symbols.
The overall procedure of \ours is also concluded in the pseudocode of Appendix~\ref{appendix_pseudocode}.

% As presented in Figure~\ref{model}, each iteration in our framework is mainly organized in three parts: (a) online exploration (Step \ding{172}-\ding{177}); (b) trajectory filtering and candidate pool updating (Step \ding{178}); and (c) data sampling and training (step \ding{179}-\ding{180}). Details will be discussed as follows. To simplify the expression, we omit the indicator of iteration $i$ in the symbols.

% sqs: 这里的核心部件，online exploration, active memory selection, ... 都黑体加粗吧，后面正好也能和subsection对的上

\subsection{Mitigating SL Data Scarcity with Online Exploration}

With the lack of well-annotated SL data for training, \ours empowers policy LLM to produce candidate symbolic solutions through autonomous interaction with the environment $\textbf{\textsc{Env}}$.
% Due to the absence of well-annotated SL data for training, the policy LLM relies on autonomous interaction with the environment $\textbf{\textsc{Env}}$ to produce candidate symbolic solutions. 
This process is named \emph{Online Exploration}, which includes three main aspects 1) self-exploration (Step \ding{172}-\ding{174}); 2) self-refinement (Step \ding{175}-\ding{177}); and 3) self-rewarding (Step \ding{178}).

\vspace{-0.75em}
\paragraph{Self-exploration}
Given the NL input $x$,
the policy model $\pi_{\theta}$ first generates $K$ diverse symbolic outputs (Step \ding{172}), formulated as $\{a_k\}_{k=1}^{K} \sim \pi_{\theta}(\cdot | x)$. These intermediate outputs will be executed in $\textbf{\textsc{Env}}$ (Step \ding{173}) to obtain the binary feedback $\{b_k\}_{k=1}^{K}$ based on $y$ (Step \ding{174}).
The above procedure allows $\pi_{\theta}$ to explore the environment autonomously and search for diverse symbolic solutions.
% The above procedure is termed as \emph{self-exploration}, where $\pi_{\theta}$ searches for potential symbolic solutions. 

\vspace{-0.75em}
\paragraph{Self-refinement.}
Considering the complexity of SL, solutions generated by the LLM may contain mistakes in symbolic format, which significantly impair the efficiency of exploration.
% As a result, we offer the self-explored symbolic solution $a_k$ as a reference for \emph{self-refinement}, resulting in the generation of solutions ${\widetilde{a}_{k=1}^{K}}$ (Step \ding{175}). 
% \emph{Self-refinement} process is formulated as $\{\widetilde{a}_k\}_{k=1}^{K} \sim \pi_{\theta}(\cdot | x;a_k)$.
% Similarly, the refined symbolic solution interacts with the environment $\textbf{\textsc{Env}}$ (Step \ding{176}) and receives the corresponding binary reward $\widetilde{b}_{k}$ (Step \ding{177}).
To address this, we utilize the above self-explored solutions $\{a_k\}_{k=1}^{K}$ as references to regenerate new refined symbolic solutions (Step \ding{175}), formulated as $\{\widetilde{a}_k\}_{k=1}^{K} \sim \pi_{\theta}(\cdot | x;a_k)$.
Similarly, these outputs will be executed in $\textbf{\textsc{Env}}$ (Step \ding{176}) and receive the corresponding binary reward $\{\widetilde{b}_k\}_{k=1}^{K}$ (Step \ding{177}).

\vspace{-0.75em}
\paragraph{Self-rewarding.}
% \zy{this part needs to be rewritten. lots of unclearness in motivation and design}
% Although binary rewards are obtained during the environment interaction process, they still fail to give general preferences for both positive and negative solutions. \zy{?}

Feedback from $\textbf{\textsc{Env}}$ merely gives the binary rewards. 
However, it remains challenging to discern preferences among various positive solutions or obtain valuable feedback from negative solutions.
% Motivated by it, we consider obtaining soft reward scores through sequence output probabilities with the following calculation:
Motivated by it, we propose a soft reward score through sequence output probabilities with the following calculation:
\vspace{-0.2cm}
\begin{equation}
    r = \overline{p}_{\theta}(a|x) = \frac{\sum\limits_{t}{\mathrm{log}} \, p_{\theta}(a_t|x;a_{<t})}{||a||},
\end{equation}

where $||a||$ is the length of the symbolic solution $a$. Based on this definition, the soft self-rewards of $a_{k}$ and $\widetilde{a}_{k}$ are derived respectively as $r_{k}$ and $\widetilde{r}_{k}$.
Considering that no extra reward model is involved, we name it \emph{self-rewarding}\footnote{\emph{Self-rewarding} step in \ours is different from~\cite{yuan2024self}, though they share the same name.} (Step \ding{178}).

% \zy{i only see two steps here, but in the figure there are 6}

\subsection{Data Selection and Training Strategies}

After the online exploration stage,
the candidate trajectories are constructed as $T_k=(x,y,a_k,b_k,r_k)$ and $\widetilde{T}_k=(x,y,\widetilde{a}_k,\widetilde{b}_k,\widetilde{r}_k)$, which are sourced from \emph{self-exploration} and \emph{self-refinement} respectively.
Next, we select premium trajectories for training the LLM $\pi_{\theta_i}$.

\vspace{-0.75em}
\paragraph{Trajectory filtering and candidate pool updating.}
To control the candidate number to maintain trajectories with high quality,
we select the superior one from $T_k$ and $\widetilde{T}_k$ to update the candidate trajectory pool (Step \ding{179}).
To facilitate automatic selection, we incorporate binary rewards and self-rewards for assessment.
Following the principle of prioritizing execution correctness, we derive the filtered trajectory $T_k^{*}$:

% \zy{why * for b and r? also it's wired to call it optimal trajectory, since there is no optimization here..}
\vspace{-0.2cm}
\begin{equation}
T_k^{*} = (x, y, a_{k}^{*}, b_{k}^{*}, r_{k}^{*}) =
\begin{cases}
(x, y, a_{k}, b_{k}, r_{k}), & \text{if } b_{k} = 1 \text{ and } \widetilde{b}_{k} = 0, \vspace{0.1em}\\
(x, y, a_{k}, b_{k}, r_{k}), & \text{if } b_{k} = \widetilde{b}_{k} \text{ and } r_{k} > \widetilde{r}_{k}, \vspace{0.1em} \\
(x, y, \widetilde{a}_{k}, \widetilde{b}_{k}, \widetilde{r}_{k}), & \text{otherwise}.
\end{cases}
\end{equation}
Notably, our filter strategy still maintains some trajectories with incorrect solutions but relatively higher rewards. These trajectories will serve as hard negative samples for the subsequent steps. 

\vspace{-0.05em}
\paragraph{Supervised fine-tuning on positive solutions.}
As we have explored diverse trajectories in $\textbf{\textsc{Env}}$, an intuitive way to bootstrap the performance of LLMs is fine-tuning with the positive solutions.
Therefore, for each input $x$, we can retrieve the positive trajectories (i.e., $b=1$) from the candidate pool.
Giving priority to more valuable solutions, we rank the trajectories in descending order based on self-rewards, resulting in the positive set $S^{+}$. 
To mitigate overfitting, we enforce a maximum of $N_1$ positive-only solutions sampled for each input $x$:

\begin{equation}
    U_{1} = \{(x, a_{m}^+) \mid m \leq \min(N_1, |S^+|) \textrm{ and } T_{m}^+ \in S^{+} \}
\end{equation}

where $m \in \mathbb{Z}^+$ means the index in the ranked set and $|\cdot|$ returns the number of trajectories in the given set. $T_{m}^{+} = (x, y, a_{m}^{+}, b_{m}^+ r_{m}^{+})$ denotes the trajectories in $S^{+}$. Following the principle of MLE, the optimized loss function can be written as:

\begin{equation}
    \mathcal{L}_{1} = -\sum\limits_{(x,a^+)\sim {U}_1}{\mathrm{log}} \, p_{\theta}(a^+|x)
\end{equation}

\vspace{-0.75em}
\paragraph{RL-free loss to learn from mistakes.}
Under the neural-symbolic setting, negative solutions may comprise a substantial portion of exploration trajectories, while also offering valuable insights for model enhancement. \ours explores motivating the policy LLM to learn from mistakes during the weak-to-strong process.
% Merely optimizing for positive solutions is inadequate, as it disregards the potential value derived from negative solutions.
% Empowered in an embodied environment, an LLM lacking self-refinement capability is inherently flawed, significantly limiting its evolution efficiency and scalability to complex cases. 
% Thus, it is crucial to motivate the policy LLM to self-refine from previous negative solutions during the weak-to-strong process.
We can obtain the ranked negative set $S^{-}$ from the candidate pool. For each input $x$, at most $N_2$ positive-negative pairs will be constructed from $S^+$ and $S^-$:

\begin{equation}
    U_{2} = \{(x, a_{m}^+, a_{m}^-) \mid m \leq \min(N_2, |S^+|-N_1, |S^-|) \textrm{ and } T_{m+|U_{1}|}^{+} \in S^{+} \textrm{ and } T_{m}^{-} \in S^{-}\}
\end{equation}

where $T_{m}^{-} = (x, y, a_{m}^{-}, b_{m}^- r_{m}^{-})$ denotes the trajectories in $S^{-}$.
Limited by the difficulty and complexity of optimizing models in an RL manner (e.g., DPO~\cite{rafailov2024direct} and PPO~\cite{schulman2017proximal}), it is challenging for reinforced-based methods such as SPIN~\cite{chen2024self} and DNO~\cite{rosset2024direct} to quickly adapt to the SL scenarios. Therefore, we design the following RL-free loss function in a contrastive manner:

\begin{equation}
    \mathcal{L}_{2} = -\sum\limits_{(x,a^+,a^-)\sim {U}_2}{\mathrm{log}}\, p_{\theta}(a^+|x;a^-)
\end{equation}

It brings two main advantages: (1) the ability of self-refinement is acquired, which benefits the scalability to complex cases; (2) compared to reinforced losses, superior training efficiency is achieved.
% It naturally inherits the efficiency of SFT, compared with reinforced losses (e.g., DPO and PPO).
Finally, the overall loss function of each iteration is simply designed as $\mathcal{L} = \mathcal{L}_{1} + \mathcal{L}_{2}$. We leave the discussion of the weighted one in the future.

\section{Experiments}

\subsection{Datasets}

We evaluate the proposed framework on three distinct domains, each with its own environment: web agents (Chrome browser), math reasoning (Python compiler), and logical reasoning (Pyke engine).
For agentic tasks, we select the widely-used web navigation benchmark MiniWob++~\citep{liu2018reinforcement}. For the math reasoning domain, we include 5 tasks: GSM8K~\cite{cobbe2021training}, MATH~\cite{hendrycks2021measuring}, GSM-Hard~\cite{gao2023pal}, SVAMP~\cite{patel2021nlp}, and AsDiv~\cite{miao2020diverse}. ProofWriter~\cite{tafjord2021proofwriter} and RuleTaker~\cite{clark2021transformers} are used to evaluate logic reasoning performance. To evaluate the generalization capability of our method, we reserve some datasets for out-of-distribution evaluation as shown in Table~\ref{task_details}. 
% All these tasks are required to be solved through interaction with the environment for execution. The training samples of the agent task are dynamically obtained from the environment for each iteration, while the other two domains receive static input data throughout training. 
Please refer to Appendix~\ref{appendix_tasks} for more details.

% \subsection{Training Setup}

\begin{table*}[t]
\centering
\footnotesize
\resizebox{\linewidth}{!}{
\begin{tabular}{l|ccccc}
    \toprule
    \textbf{Domains} &\textbf{Held-in Tasks} &\textbf{Held-out Tasks} &\textbf{\#Samples} & \textbf{Static ?} &\textbf{Env.} \\
    \midrule
    Web Agent & MiniWob++ & - &2,200 &No &Chrome browser\\
    Math Reasoning & GSM8K, MATH &GSM-H, SVAMP, AsDiv &13,492 &Yes &Python compiler \\
    Logic Reasoning & ProofWriter & RuleTaker &3,600 &Yes &Pyke engine \\
    \bottomrule
\end{tabular}}
\caption{Details and statistics of evaluated domains. \emph{\#Samples} denotes the number of input samples per iteration. \emph{Static?} indicates whether the input data remains the same across all iterations. \emph{Env.} column presents the external solver for execution.}
\vspace{-1.25em}
\label{task_details}
\end{table*}

\subsection{Baselines and Training Details}
\label{sec:base_and_train}

Following the categorization of Figure~\ref{paradigm}, we consider the respective three lines of baselines. All baselines are reproduced under the same codebase for a fair comparison.

\textbf{Distill-then-Finetune.}
GPT-4 and Claude-2 are selected as strong teacher LLMs in this approach. By prompting teacher LLMs, we obtain the symbolic trajectories with correct answers to fine-tune the LLM. In light of the high time and financial cost of calling API, each input will be prompted only once.

\textbf{Reinforced Self-Training.}
We implement two RL-based self-training baselines: \emph{Self-Rewarding}~\cite{yuan2024self} and \emph{iterative SFT+DPO}. For the former, we follow the official implementation to first warm up the weak LLM using human annotation from OpenAssistant~\cite{kopf2024openassistant}. The latter is a simple variation of \ours that mainly separates the training into two stages, with positive solutions for SFT and positive-negative pairs for DPO.
% We implement two self-training baselines with RL-based methods. 
% The first one is \emph{Self-Rewarding}~\cite{yuan2024self}. Following the original implementation, we employ human-ranked response data provided in OpenAssistant~\cite{kopf2024openassistant} as the extra training samples to first warm up the weak LLM. Considering the performance and training cost, the iteration number of \emph{self-rewarding} is fixed to 4.
% The second one is named \emph{iterative SFT+DPO}, where the weak LLM will sequentially go through SFT, sample generation and DPO stages in one iteration.

\textbf{Env-guided Self-Training.}
Since there are no existing baselines for this approach, we consider extending the 
NL-centric self-training method STaR~\cite{zelikman2022star} to support neural-symbolic scenarios. It is worth noting that STaR only uses positive samples for behavior cloning. 
For the methods under this paradigm (including \ours) we optimize LLM from scratch in each iteration with the updated training samples.

% Apart from our proposed framework, we implement one additional baseline \emph{STaR+Env.} to illustrate the superiority of 
% \emph{Env-guided Self-Training} paradigm. Since previous work \emph{STaR}~\cite{zelikman2022star} can not directly generalize to neural-symbolic scenarios (e.g., web agent tasks), we combine it with the environment interaction in our paradigm.

% \zy{looks like one missing baseline is the reward model based approach. can self-rewarding fit into this line of baselines? Also, what is iterative sft+dpo} \zy{the design of baselines does not match figure 1. Following figure 1, you should have three type of baselines distill reward-based and other self-training baselines. } \xfz{\textbf{self-rewarding is reinforced self-training method} with DPO as the loss and warm-up training for reward generation (LLM itself as the reward model).....\textbf{iterative SFT+DPO is also reinforced self-training}, which mainly replaces L2 with DPO-style loss (as DPO claims: LLM can be an implicitly RM with DPO design)......STaR+Env. can be categorized into other self-training methods}

Except for \emph{Distill-then-Finetune} baselines, all other methods, ours included, utilize few-shot prompting to acquire training samples as a cold start. The few-shot numbers for the web agent, math reasoning, and logic reasoning domains are set to 1, 3, and 1 respectively. We also include few-shot results on weak LLM for comparison. For a fair evaluation, all baselines are optimized to generate symbolic outputs (e.g., Python code) rather than natural language outputs, following PoT style~\cite{chen2022program}. Please refer to Appendix~\ref{appendix_implementation} for other details.

We use LLaMA2-Chat 7B/13B models for the evaluation.
% Generalization to other base models (e.g., DeepSeek-Chat~\cite{deepseek-llm}) will also be discussed in the analysis. 
At each generation step (i.e., Step \ding{172},\ding{175}), the candidate size $K$ is set to 5. 
% To ensure the efficiency of the framework, we employ vLLM~\cite{kwon2023efficient} to speed up the generation. For model training of each iteration, we adopt the full finetuning strategy accelerated by Deepspeed Zero3 and FlashAttention2.
The total iteration number for web agent, math, and logic tasks is set to 5, 10, and 8 respectively, unless otherwise stated.
For each input, $N_1$ and $N_2$ are fixed to 10 and 2 respectively.
All the self-training experiments are implemented on 8*A100 of 80GB VRAM.
Please refer to Appendix~\ref{appendix_trainingdetails} for other details.

\begin{table*}[t]
\centering
\footnotesize
\resizebox{\linewidth}{!}{
\begin{tabular}{l|c|ccccc|cc|c}
    \toprule
    \multirow{2}{*}{\textbf{Models}}& \multicolumn{1}{c|}{\textbf{Agent}} &\multicolumn{5}{c|}{\textbf{Math Reasoning}} &\multicolumn{2}{c|}{\textbf{Logical Reasoning}} &\multirow{2}{*}{\textbf{Avg.}} \\
    &\textbf{MiniWob++} &\textbf{GSM8K} & \textbf{MATH} & \textbf{GSM-H} &\textbf{SVAMP} &\textbf{ASDiv} &\textbf{ProofWriter} &\textbf{RuleTaker} & \\
    \midrule
    \multicolumn{1}{l|}{\textbf{Is Held-out ?}} &\color{red}\usym{2717} &\color{red}\usym{2717}  &\color{red}\usym{2717} &\color{ao(english)}\checkmark
    &\color{ao(english)}\checkmark &\color{ao(english)}\checkmark &\color{red}\usym{2717}  &\color{ao(english)}\checkmark & \\
    \midrule
    \multicolumn{10}{c}{\cellcolor{gray!25} LLaMA2-Chat (7B)} \\
    \midrule
    LLaMA2-Chat (few-shot) &51.14 &12.21 &1.32 &10.69 &22.00 &25.86 &34.83 &47.44 &25.69 \\
    \midrule
    \textbf{Distill-then-Finetune} & & & & & & & & & \\
    GPT-4 + LLaMA2-Chat &81.14 &53.07 &18.84 &47.84 &66.80 &68.75 &34.33 &48.88 &52.46\\
    Claude-2 + LLaMA2-Chat &82.80 &52.69 &18.17 &44.88 &70.50 &69.85 &36.17 &49.17 &53.03 \\
    \midrule
    \textbf{Reinforced Self-Training} & & & & & & & & & \\
    Self-Rewarding &69.39 &40.03 &10.70 &31.69 &58.20 &61.55 &32.17 &50.04 &44.22 \\
    Iterative SFT+DPO &77.05 &54.81 &14.75 &47.08 &70.10 &66.22 &49.00 &58.82 & 54.73 \\
    \midrule
    \textbf{Env-guided Self-Training} & & & & & & & & & \\
    STaR + Env. &83.71 &58.23 &15.97 &46.63 &67.50 &68.46 &50.17 &58.60 &55.91 \\
    \ours &\textbf{85.38} &\textbf{58.98} &\textbf{19.00} &\textbf{48.52} &\textbf{72.40} &\textbf{69.80} &\textbf{52.83} &\textbf{62.63} &\textbf{58.69}  \\
    \midrule
    \multicolumn{10}{c}{\cellcolor{gray!25} LLaMA2-Chat (13B)} \\
    \midrule
    LLaMA2-Chat (few-shot) &60.00 &34.87 &6.07 &28.96 &45.00 &46.61 &35.83 &51.50 &38.61 \\
    \midrule
    \textbf{Distill-then-Finetune} & & & & & & & & & \\
    GPT-4 + LLaMA2-Chat &80.15 &62.85 &23.64 &53.98 &73.00 &73.52 &34.17 &50.61 &56.49 \\
    Claude-2 + LLaMA2-Chat &84.77 &62.24 &23.47 &52.08 &76.30 &74.05 &36.00 &48.45 &57.17 \\
    \midrule
    \textbf{Reinforced Self-Training} & & & & & & & & & \\
    Self-Rewarding &74.55 &50.80 &13.97 &41.24 &74.10 &71.37 &37.33 &56.66 &52.50 \\
    Iterative SFT+DPO &82.73 &63.84 &22.32 &50.57 &77.30 &70.94 &51.00 &59.47 &59.77 \\
    \midrule
    \textbf{Env-guided Self-Training} & & & & & & & & & \\
    STaR + Env. &85.15 &63.61 &20.57 &53.37 &74.70 &74.76 &52.33 &60.33 &60.60 \\
    \ours &\textbf{87.12} &\textbf{68.31} &\textbf{26.04} &\textbf{57.54} &\textbf{78.30} &\textbf{75.52} &\textbf{54.83} &\textbf{60.84} &\textbf{63.56}  \\
    \bottomrule
\end{tabular}
}
\caption{Main Results on Agent, Math Reasoning and Logical Reasoning domain. Notably, we report the average performance across extensive tasks in MiniWob++ benchmark (refer to Appendix~\ref{appendix_miniwob} for details). \emph{Is Held-out?} row distinguishes the held-in and held-out tasks. \emph{Avg.} column reports the averaged performances on all tasks.}
\label{exp_main}
\vspace{-0.9em}
\end{table*}

\subsection{Main Results}
Table~\ref{exp_main} presents the evaluation results. We include an LLaMA2-Chat (few-shot) baseline for reference purposes, all other methods are tested under the zero-shot setting.
% the main results on held-in and held-out tasks across diverse domains, compared with some strong baselines. Apart from LLaMA2-Chat (few-shot), other methods are tested under the zero-shot setting. \ckz{The order of paradigm and framework advantages?}

\vspace{-0.75em}
\paragraph{\ours presents consistent superiority over strong baselines.} 
Evolving from LLaMA2-Chat, 
\ours significantly enhances average performances by 30.00\% and 24.95\% for the 7B and 13B variants, respectively. Compared with \emph{Distill-then-Finetune} methods, 5.66\%-7.13\% improvements are obtained.
In addition to its superior performance, \ours offers greater scalability compared with the significant costs associated with using strong LLMs (e.g., GPT-4).
Besides it, the obvious advantages over \emph{Reinforced Self-Training} and other \emph{Env-guided Self-Training} methods are observed, with 2.78\%-14.47\% average gains.
The combination of the performance as well as the training efficiency makes \ours stand out among these strong baselines.

\vspace{-0.75em}
\paragraph{Env-guided Self-Training exhibits strong scalability to neural-symbolic scenarios.}
Compared to the other two approaches, 
\emph{Env-guided Self-Training} is more applicable to complex neural-symbolic scenarios, especially in agentic tasks where NL-centric methods inherently exhibit limitations.
Besides the great performances of \ours, previous methods \emph{STaR} can also benefit from the supervision signals acquired in $\textbf{\textsc{Env}}$, which helps the evolution progress.

\subsection{Evolution Progress for Self-Training Frameworks}

\begin{figure}[t]
    \centering
    \begin{subfigure}[b]{0.99\textwidth}
        \includegraphics[width=\textwidth, keepaspectratio]{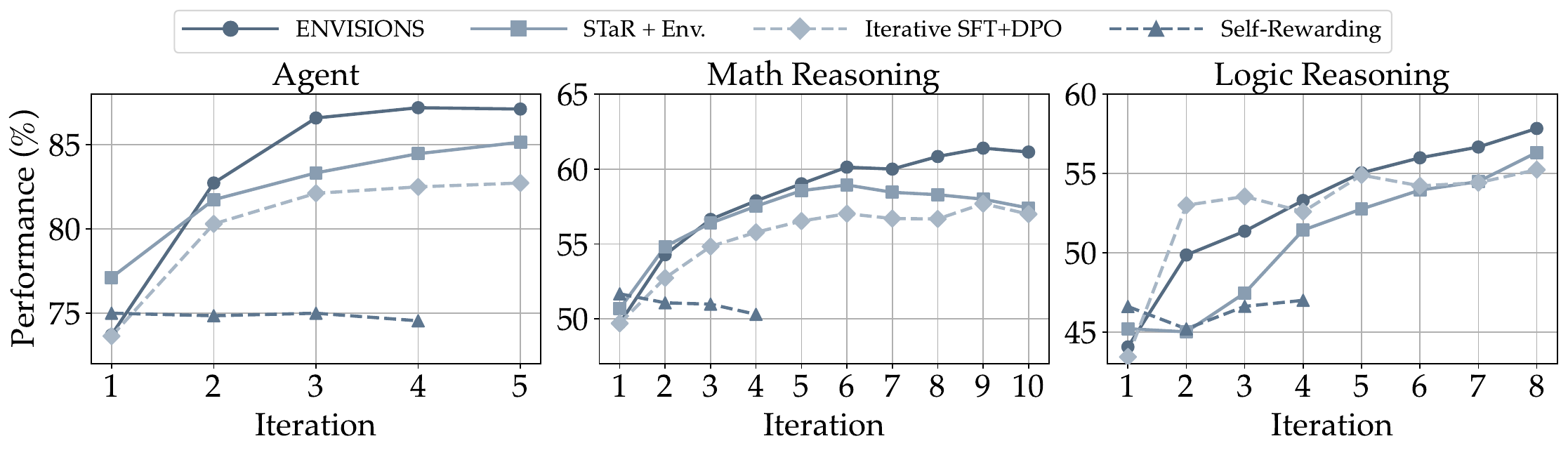}
    \end{subfigure}
    \caption{Performance evolution of self-training methods on LLaMA2-Chat 13B model. \emph{Reinforced Self-Training} approaches are represented by dashed lines, while \emph{Env-guided} ones are in solid lines.}
    \label{performance_curve}
    \vspace{-1.25em}
\end{figure}

In Figure~\ref{performance_curve}, we present the iterative evolution curves of the self-training frameworks with LLaMA2-Chat (13B) as the LLM, which clearly shows the procedure of weak-to-strong transformation. Limited by space, we leave the discussion on the evolution of both performance and explored sample numbers with the 7B version in Appendix~\ref{appendix_curves}.

% \vspace{-0.75em}
\paragraph{\ours combines high evolutionary efficiency and sustainability.}
In the initial iterations, \ours demonstrates swift adaptability to different scenarios. This indicates that exceptional performance can be achieved with minimal time for data collection in \ours. Additionally, \ours stands out as a more sustainable option when compared to other baselines. For instance, in math reasoning tasks of Fig.~\ref{performance_curve}(b), all baseline methods achieve saturated performance levels by $6^{th}$ iteration. However, our framework continues to exhibit evolutionary progress.

\vspace{-0.75em}
\paragraph{Reinforced baselines are largely flawed during iterations.}
The incorporation of reinforced loss (e.g., DPO) brings difficulty in optimization and greatly restricts the evolutionary scales of the LLM to adapt to the neural-symbolic scenarios.
\emph{Self-Rewarding} exhibits largely reduced benefits during iterations, in contrast to its impressive performance in NL-centric tasks. 
% This observation highlights the distinctive challenges \ckz{what is the distinctive challenges} encountered in complex neural-symbolic scenarios. 
For \emph{Iterative SFT+DPO}, the SFT stage boosts the ability in effective exploration. However, the subsequent DPO stage imposes a slight improvement while significantly reduce the training efficiency.

\subsection{Generalization to Various Backbones}

To demonstrate the generalizability,
we apply 
% \sqs{our method} 
\ours
to enhance two additional base LLMs on mathematical reasoning tasks:
(1) DeepSeek-Chat~\cite{deepseek-llm} model of 7B size,
% The first LLM is  DeepSeek-Chat of 7B version~\cite{deepseek-llm}, which is recognized as a powerful contender as the base LLM. 
% The second model, 
% Llemma~\cite{azerbayev2023llemma}, is a highly capable domain-specific LLM optimized for math reasoning. 
\begin{wrapfigure}{r}{7.35cm}
\vspace{-0.85em}
    \centering
    \includegraphics[width=\linewidth]{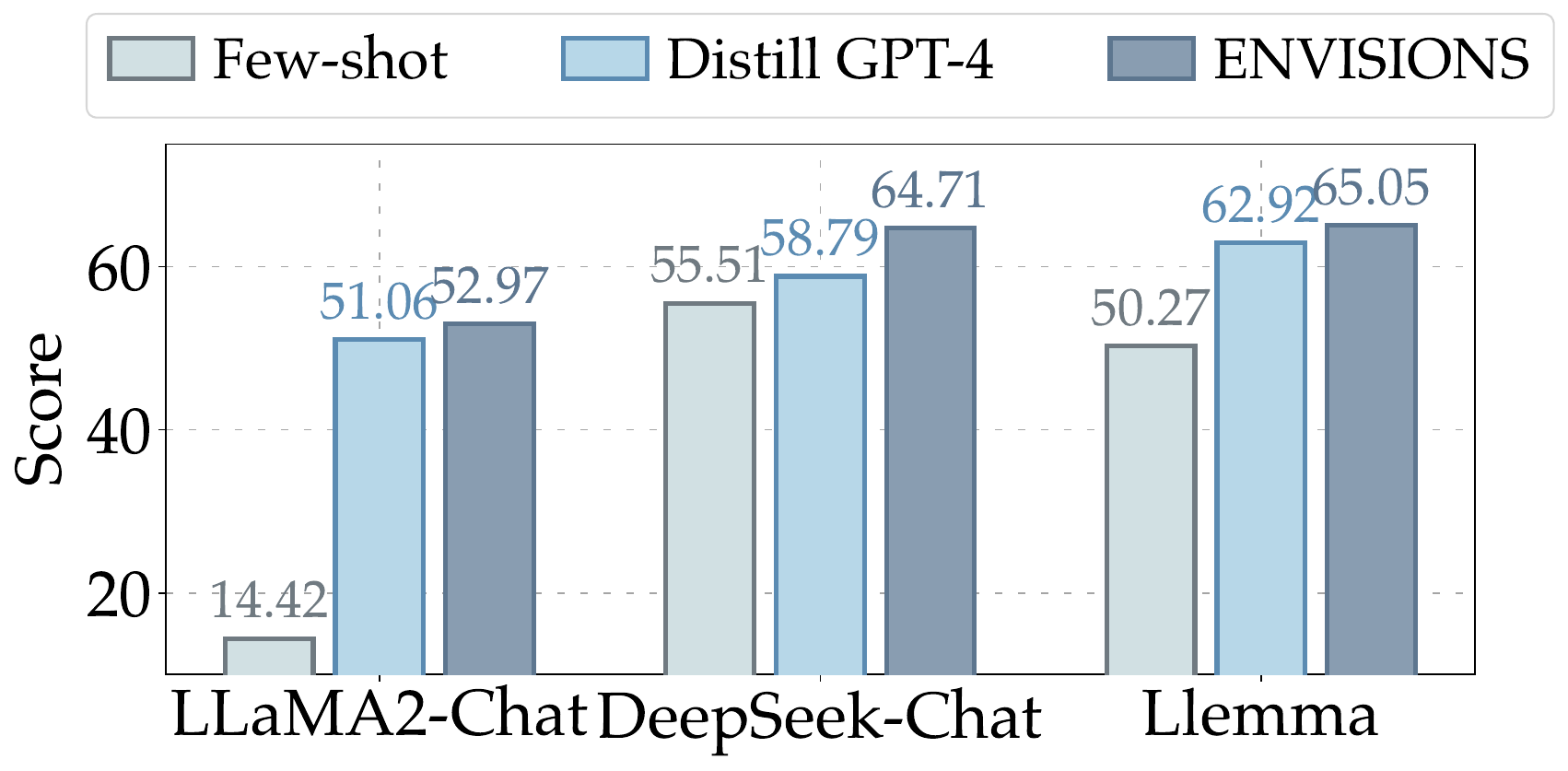}
    \vspace{-1.5em}
    \caption{Generalization to different LLMs. The performances on math reasoning tasks are reported.}
    \label{fig:various_base}
    \vspace{-2em}
\end{wrapfigure}
which is a foundational LLM and (2) Llemma~\cite{azerbayev2023llemma},
a competent domain-specific LLM optimized for math reasoning.
Figure~\ref{fig:various_base} shows the comparisons with \emph{Few-shot Prompting} and \emph{Distill GPT4-then-Finetune}.
It is observed that our framework still works for strong foundation LLMs, with 9.20\% and 14.78\% performance boosts for DeepSeek-Chat and Llemma respectively. This demonstrates that our framework can not only convert LLMs from weak to strong, but also elevate LLMs from strong to stronger.
\section{Analysis}
This section will make an in-depth analysis of the underlying reason behind \ours's success. 
% discussion on 1) the impacts of key components in \ours, 2) novel insights and analysis on the superiority of \ours especially compared with reinforced methods.

\begin{table*}[t]
\centering
\footnotesize
\resizebox{\linewidth}{!}{
\begin{tabular}{l|c|ccccc|cc|c}
    \toprule
    \multirow{2}{*}{\textbf{Models}}& \multicolumn{1}{c|}{\textbf{Agent}} &\multicolumn{5}{c|}{\textbf{Math Reasoning}} &\multicolumn{2}{c|}{\textbf{Logical Reasoning}} &\multirow{2}{*}{\textbf{Avg.}} \\
    &\textbf{MiniWob++} &\textbf{GSM8K} & \textbf{MATH} & \textbf{GSM-H} &\textbf{SVAMP} &\textbf{ASDiv} &\textbf{ProofWriter} &\textbf{RuleTaker} & \\
    % \midrule
    % \multicolumn{1}{l|}{\textbf{Is OOD Setting ?}} &\color{red}\usym{2717} &\color{red}\usym{2717}  &\color{red}\usym{2717} &\color{ao(english)}\checkmark
    % &\color{ao(english)}\checkmark &\color{ao(english)}\checkmark &\color{red}\usym{2717}  &\color{ao(english)}\checkmark & \\
    \midrule
    \multicolumn{10}{c}{\cellcolor{gray!25} LLaMA-2-Chat (7B)} \\
    \midrule
    \ours &\textbf{85.38} &\textbf{58.98} &\textbf{19.00} &\textbf{48.52} &\textbf{72.40} &\textbf{69.80} &\textbf{52.83} &\textbf{62.63} &\textbf{58.69}  \\
    \quad w/o self-refine &84.92 &56.86 &18.20 &48.14 &68.70 &67.89 &42.00 &58.60 &55.66 \\
    \quad w/o self-reward &84.47 &58.61 &18.75 &47.92 &71.10 &68.46 &47.33 &59.61 &57.03 \\
    \quad w/o candidate pool &83.86 &57.77 &17.55 &47.16 &70.90 &68.03 &49.17 &59.18 &56.70 \\
    \quad w/o $\mathcal{L}_{2}$ loss &81.89 &55.88 &18.90 &47.16 &67.60 &67.75 &47.67 &57.88 &55.59 \\
    \midrule
    \multicolumn{10}{c}{\cellcolor{gray!25} LLaMA-2-Chat (13B)} \\
    \midrule
    \ours &\textbf{87.12} &\textbf{68.31} &\textbf{26.04} &\textbf{57.54} &\textbf{78.30} &\textbf{75.52} &\textbf{54.83} &\textbf{60.84} &\textbf{63.56}  \\
    \quad w/o self-refine &84.24 &65.96 &24.95 &55.34 &77.70 &73.90 &51.00 &57.59 &61.34 \\
    \quad w/o self-reward &85.45 &67.02 &25.59 &55.57 &77.80 &74.05 &51.50 &60.69 &62.21 \\
    \quad w/o candidate pool &85.61 &66.89 &24.19 &53.07 &77.20 &72.90 &51.33 &58.96 &61.27 \\
    \quad w/o $\mathcal{L}_{2}$ loss &81.59 &63.08 &20.00 &51.18 &74.30 &71.23 &50.33 &60.19 &58.99 \\
    \bottomrule
\end{tabular}
}
\caption{Ablation studies on key components.}
\label{ablate}
\end{table*}

\subsection{What is the Impact of Key Components in \ours?}

Some key components are ablated independently to verify their effectiveness in Table~\ref{ablate}. \emph{w/o self-refine} ablates both the self-refinement process (i.e., Step \ding{175}-\ding{177}) and $\mathcal{L}_{2}$. \emph{w/o self-rewards} replaces the trajectory ranking on the self-rewarding strategy with random sampling. \emph{w/o long-term memory} only utilizes the generated trajectories from the current iteration for training. \emph{w/o $\mathcal{L}_{2}$ loss} ablates the optimization with positive-negative trajectory pairs. 

Of these components, self-refine-oriented optimizations (i.e., self-refinement and $\mathcal{L}_{2}$ loss) play key roles in boosting the performances.
As one of the key contributions, the design of $\mathcal{L}_{2}$ loss leads to 3.10\%-4.57\% improvements in \ours.
It makes full use of negative trajectories while maintaining training efficiency in an RL-free style. Especially in agent tasks, \ours benefits a lot from $\mathcal{L}_{2}$ loss, with 3.49\%-5.53\% performance gains.

\begin{figure}[t]
    \centering
    \begin{subfigure}[b]{0.99\textwidth}
        \includegraphics[width=\textwidth, keepaspectratio]{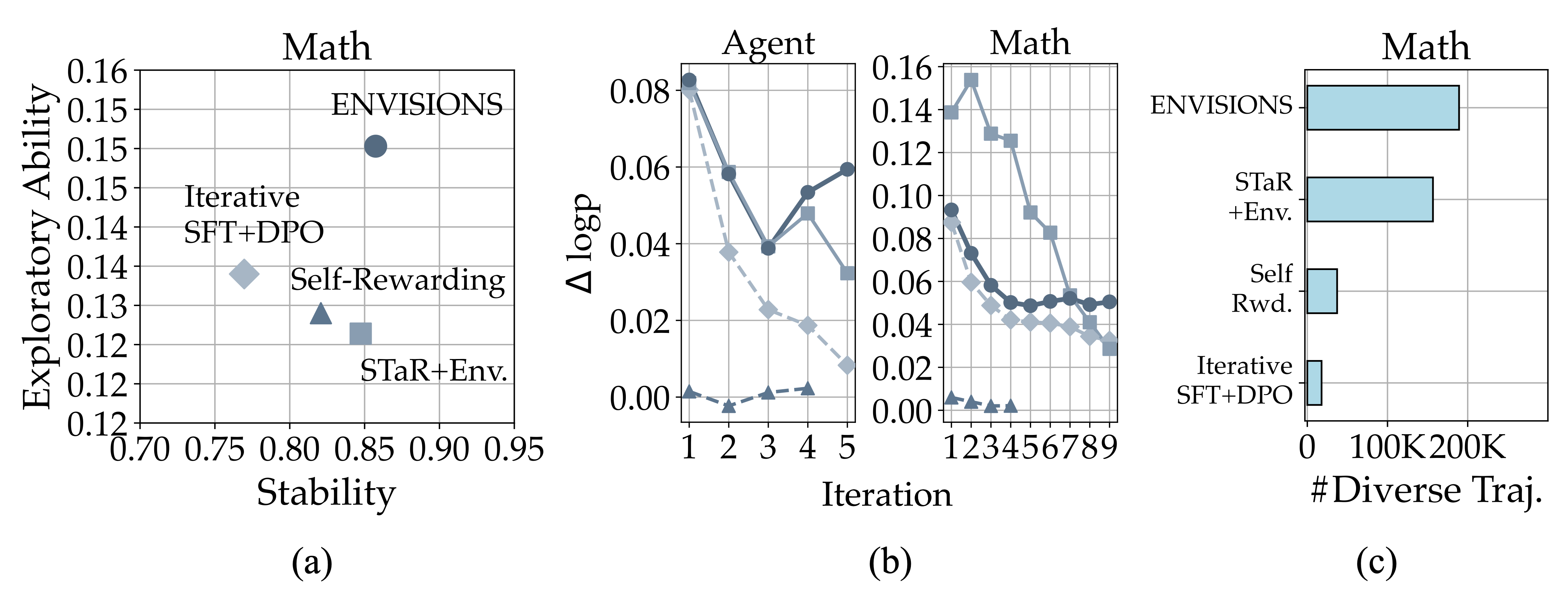}
        % \caption{LLaMA2-Chat (7B)}
    \end{subfigure}
    % \vspace{-1em}
    \vskip -0.75em % 调整这个值来减少子图之间的间隙
    \begin{subfigure}[b]{0.99\textwidth}
        \includegraphics[width=\textwidth, keepaspectratio]{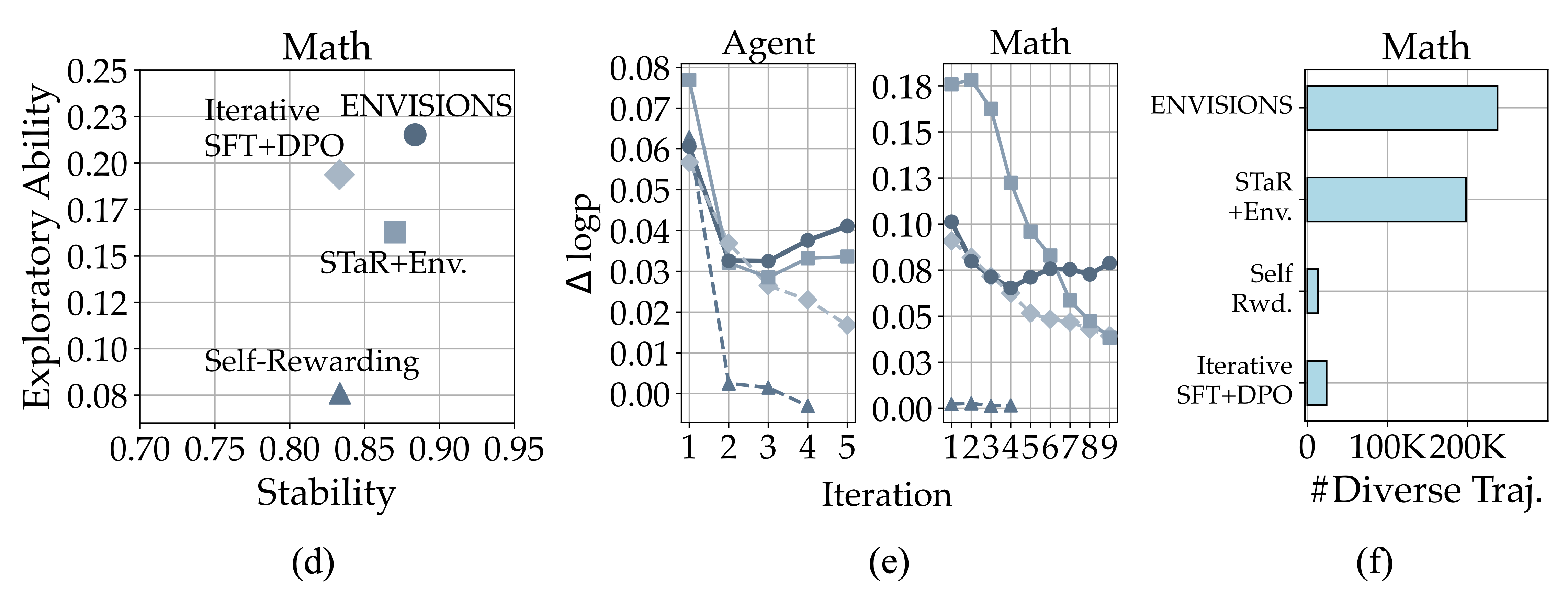}
        % \caption{LLaMA2-Chat (7B)}
    \end{subfigure}
    \vspace{-0.5em}
    \caption{In-depth analysis from three perspectives. The first row (i.e., (a),(b),(c)) and the second row (i.e., (d),(e),(f)) represent the results on LLaMA2-Chat (7B) and LLaMA2-Chat (13B) respectively.
    Plots in fig.(b),(e) correspond to the methods represented by the same colors in fig.(a),(d).
    % The line colors in the line chart correspond to the methods represented by the scatter plot.
    }
    \label{analysis}
    \vspace{-1.25em}
\end{figure}

\subsection{What is Behind the Superiority of \ours?}

We provide in-depth evidence and analysis on the superiority of \ours from three distinctive views: (1) exploratory ability and stability; (2) log probability margin between positive and negative solutions; and (3) diversity of synthetic samples.

\vspace{-0.75em}
\paragraph{Balanced exploratory ability and stability are key to success in weak-to-strong.}
To effectively navigate the environment and acquire new skills autonomously, two factors are crucial: 1) promptly resolving extensive samples to collect correct trajectories, and 2) minimizing the potential loss of knowledge from previously-solved samples.
We employ two metrics \emph{exploratory ability} and \emph{stability} to evaluate the LLM (both of them are the higher, the better). Refer to Appendix~\ref{appendix_definition} for definition details. In Figure~\ref{analysis}(a)(d), \ours demonstrates remarkable performance in achieving a balance between exploratory ability and stability. By leveraging the candidate trajectory pool and self-rewards, \ours effectively retains high-quality positive solutions during training, significantly mitigating the issue of forgetting previous trajectories. Additionally, the RL-free loss $\mathcal{L}_2$ enables flexible updates of the LLM, enhancing its exploration capabilities.

% We use two metrics to evaluate the LLM performance during self-training: \emph{exploratory ability} and \emph{stability}. The former assesses how well the LLM can navigate and solve new cases, while the latter measures its ability to consistently perform well on previously solved cases. The implementations refer to Appendix\xfz{add details}. In Figure~\ref{analysis}(a), our framework demonstrates superior performance in achieving a balance between exploratory ability and stability \ckz{why? which design in the framework leads to such advantages?}. In contrast, the reinforced baselines fall behind in this regard.

\vspace{-0.75em}
\paragraph{Clearly distinguishing positive and negative solutions can help the LLM optimization.}
During the optimization process, it is inevitable for the log probability of both positive and negative trajectories to increase simultaneously~\cite{hong2024reference}.
However, clearly keeping the probability margins ($\Delta \mathrm{logp}$) between positive-negative pairs is crucial to facilitate the optimization. Fig.~\ref{analysis}(b)(e) shows the analysis of $\Delta \mathrm{logp}$ during iterations. It is observed \ours keeps the margin within a reasonable range, while reinforced methods exhibit a rapid decrease in $\Delta \mathrm{logp}$. It indicates the unsuitablity of DPO to the exploration setting and the importance of feedback from $\textbf{\textsc{Env}}$. Notably, \emph{STaR+Env.} fails to keep the stable margins in the math domain, since it merely utilizes positive data for training, which fails to distinguish negative ones and leads to overfiting on the limited number of solutions. Such finding corresponds to the lack of exploratory ability in Fig.~\ref{analysis}(a)(d).

% \ckz{why only three lines in fig4(b)} Figure~\ref{analysis}(b) illustrates the analysis of the log probability disparity ($\Delta \mathrm{logp}$) between positive and negative trajectories. It is observed that reinforced\sqs{RL-based} methods exhibit a rapid decrease in $\Delta \mathrm{logp}$. Since the \emph{Self-Rewarding} approach lacks clear guidance from the environment, it faces challenges in effectively distinguishing between preferences.

% \vspace{-0.75em}
\paragraph{Diverse trajectories are what you need for self-training.}
In Fig.~\ref{analysis}(c)(f), we compare the number of correct and unique trajectories by the last iteration. It demonstrates the huge strengths of \ours in synthesizing diverse trajectories. It largely surpasses \emph{Reinforced Self-Training} approaches, which is one of the underlying reasons for our superiority. In fact, the LLM updates in RL methods are restricted by \emph{KL} constraints, which ultimately impact the diversity of the generated trajectories. Moreover, \emph{Distill GPT-4} and \emph{Distill Claude2} lead to 10,831 and 8,561 diverse trajectories with one iteration. Since repeatedly calling strong LLMs involves extremely high cost and cumbersome prompt optimizations, they are far from sustainable compared with \ours. 
% We discuss the cost comparison in Appendix \xfz{add} to verify the efficiency of \ours. 

% Reinforced\sqs{RL-based} methods exhibit significant weakness in producing diverse trajectories. 
% One of the reasons\sqs{contributing factors} is the restrictive nature of the \emph{KL} constraint term in the reinforced loss, 
% which greatly limits the updates of LLM\sqs{unnecessary?}. 
% Notably, our framework stands out\sqs{distinguish itself} among other self-training methods due to its exceptional ability in data synthesis.

\subsection{How does the Training Recipe Matter in Iterative Self-Exploration?}

\begin{wraptable}{r}{0.585\textwidth}
\centering
\footnotesize
\begin{tabular}{l|ccc}
    \toprule
    \multirow{1}{*}{\textbf{Tasks}}& \textbf{Cont.} &\textbf{\ours} &$\Delta$ \\
    \midrule
    \multicolumn{4}{c}{\cellcolor{gray!25} LLaMA-2-Chat (7B)} \\
    \midrule
    Agent &78.18 & 85.38 &\color{ao(english)}{+7.20} \\
    Math Reasoning &51.20 &53.74 &\color{ao(english)}{+2.54} \\
    Logic Reasoning &46.20 &57.73 &\color{ao(english)}{+11.53} \\
    \midrule
    Average &53.32 &58.69 &\color{ao(english)}{+5.37} \\
    \bottomrule
\end{tabular}
\caption{Comparisons between training strategies. \emph{Cont.} column denotes the performances of \ours under the continual training setting.}
\label{exp_cont}
\vspace{-1.5em}
\end{wraptable}

In each iteration of \ours, we optimize the policy LLM from scratch (e.g., LLaMA2-Chat) with the updated training trajectories.
Such a training recipe is expected to bring stability to the training process, compared with the strategy of continuous training based on previous checkpoints.
Table.~\ref{exp_cont} presents the performance comparisons.
Obvious superiority of \ours is observed across these three domains, with an average improvement of 5.37\%.
Training from previous checkpoints does affect the exploration.
For the RL-based self-training method, the training of the policy LLM is constrained within the range of the reference model by the KL term.
In order to enable continuous evolution, the policy LLM is required to be updated from the checkpoint of the previous iteration.
It is also one of the main causes of their sub-optimal performances.

% \begin{table*}[t]
% \centering
% \footnotesize
% \resizebox{0.5\linewidth}{!}{
% \begin{tabular}{l|ccc}
%     \toprule
%     \multirow{1}{*}{\textbf{Tasks}}& \textbf{Cont.} &\textbf{\ours} &$\Delta$ \\
%     \midrule
%     \multicolumn{4}{c}{\cellcolor{gray!25} LLaMA-2-Chat (7B)} \\
%     \midrule
%     Agent &78.18 & 85.38 &\color{ao(english)}{+7.20} \\
%     Math Reasoning &51.20 &53.74 &\color{ao(english)}{+2.54} \\
%     Logic Reasoning &46.20 &57.73 &\color{ao(english)}{+11.53} \\
%     \midrule
%     Average &53.32 &58.69 &\color{ao(english)}{+5.37} \\
%     % \midrule
%     % \multicolumn{4}{c}{\cellcolor{gray!25} LLaMA-2-Chat (13B)} \\
%     % \midrule
%     % Agent &80.76 &87.12 &\color{ao(english)}{+6.36} \\
%     % Math Reasoning & &61.14 & \\
%     % Logic Reasoning &49.62 &57.84 &\color{ao(english)}{+8.22} \\
%     % \midrule
%     % Average & &63.56 \\
%     \bottomrule
% \end{tabular}
% }
% \caption{Comparisons between training strategies. \emph{Cont.} column denotes the performances of \ours under the continual training setting.}
% \label{exp_cont}
% \end{table*}

% \vspace{-0.4cm}
\section{Conclusion}
\vspace{-0.3cm}
This paper focuses on converting LLMs from weak to strong in increasingly promising neural-symbolic scenarios, without human-annotated symbolic training data. In view of two key challenges, i.e., 1) the scarcity of symbolic training data, and 2) the inherent weakness of LLMs in addressing SL, we conclude the env-guided self-training approach. Built on it, we propose a novel neural-symbolic self-training framework \ours. 
% To collect symbolic trajectories, \ours conducts online exploration through self-exploration, self-refinement and self-rewarding. Then an RL-free contrastive loss is designed 
Extensive experiments across three domains verify the remarkable performances.
In-depth analysis on the superiority of \ours from three distinctive views provide novel insights for future researches.

\bibliography{custom}
\bibliographystyle{unsrtnat}

\clearpage

\newpage

\appendix

\iffalse
\section{Limitations and Broader Impacts}
\label{sec:lim}

\paragraph{Limitations.}
% Due to the reliance on the feedback from environments,
% the proposed VISIONS framework requires pre-configuration of environments for different tasks (e.g., agent tasks), 
% making it somewhat more complex than traditional self-training methods.
% Additionally, since VISIONS employs complete trajectories for training, its computational overhead is higher than that of self-training methods which solely depend on model outputs.

Our limitations are twofold: 
(1) We have demonstrated the efficacy and superiority of \ours on a wide range of tasks across three domains. However, there remain further domains worthy of exploration (e.g., environments with visual information, robotic controlling and planning), 
They are also great importance, which we leave for future work. 
(2) Constrained by computational resources, our experiments were primarily conducted on 7B and 13B models. Therefore, it remains questionable that whether \ours can be scalable to larger size. 
% We anticipate applying this framework to larger models in the future.

\paragraph{Broader Impacts.} This paper presents work whose goal is to advance the field of deep learning and neuron-symbolic research. We do not anticipate it will cause negative societal impacts, such as potential malicious or unintended uses.
% We do not anticipate 

\fi

\section{Implementation Details}
\label{appendix_implementation}
In this section, we provide some details of the implementation.

\subsection{Training Details}
\label{appendix_trainingdetails}
The SFT training in both our framework and baselines is conducted on 8*A100 with a maximum length of 2,048. They are optimized and accelerated with Deepspeed Zero3 and FlashAttention2. The AdamW optimizer~\cite{loshchilov2017decoupled} is leveraged with a \emph{Linear} learning rate of 2e-5. The SFT training epoch number of each iteration is set to 2, 1, 2 for agent, math reasoning, and logic reasoning tasks respectively.

For the DPO training stage in baseline methods, it is also conducted on 8*A100 with a maximum length of 2,048. The \emph{Linear} learning rate is 5e-7 with a warm-up ratio of 0.1. The epoch number for each domain is the same as the SFT stage.

\subsection{Test Tasks and Benchmark}
\label{appendix_tasks}
The experiments in the main paper primarily cover three domains: web agent, math reasoning, and logic reasoning. We have concluded some key details in Table~\ref{task_details}. In Table~\ref{appendix_task}, we attach extra information on the test tasks and benchmark.

\begin{table*}[h]
\centering
\footnotesize
\resizebox{\linewidth}{!}{
\begin{tabular}{l|cccccc}
    \toprule
    \textbf{Domains} &\textbf{Task name} &\textbf{Is Held-out?} &\textbf{\#Test Samples} &\textbf{Beam Size} &\textbf{Max. Length} &\textbf{Sources} \\
    \midrule
    \multirow{1}{*}{Web Agent} & MiniWob++ & &30 ($\times$44 tasks) &1 &2,048 &~\citet{liu2018reinforcement} \\
    \midrule
    \multirow{5}{*}{Math Reasoning} & GSM8K & &1,319 &2 &2,048 &\citet{cobbe2021training} \\
    &MATH & &4,001 &2 &2,048 &\citet{hendrycks2021measuring} \\
    &GSM-Hard & \checkmark &1,319 &2 &2,048 &\citet{gao2023pal} \\
    &SVAMP & \checkmark &1,000 &2 &2,048 &\citet{patel2021nlp} \\
    &AsDiv & \checkmark &2,096 &2 &2,048 &\citet{miao2020diverse} \\
    \midrule
    \multirow{2}{*}{Logic Reasoning} & ProofWriter & &600 &1 &4,096 &\citet{tafjord2021proofwriter} \\
    & RuleTaker & \checkmark &1,389 &1 &4,096 &\citet{clark2021transformers}\\
    \bottomrule
\end{tabular}}
\caption{Details of test tasks and benchmarks.}
\label{appendix_task}
\end{table*}

Unless otherwise stated, all these test tasks are evaluated under the zero-shot setting. For MiniWob++ benchmark, we select 44 tasks for the test~\cite{cheng2024seeclick}, each with 30 randomly generated samples. All the above settings are consistent among all baseline methods.

\section{Pseudocode of \ours}
\label{appendix_pseudocode}
The self-training framework \ours can be expressed through the following pseudocode.

\begin{algorithm}[ht]
  \SetAlgoLined
  \KwIn{Data pair $\{(x,y)\}$, environment $\textbf{\textsc{Env}}$, candidate trajectory pool \textsc{POOL}, weak LLM $\pi_{\theta_0}$, number of generated samples $K$, number of iteration $T$.}
  \KwOut{Strong LLM $\pi_{\theta}^{*}$.}
  \tcp{Initialize}
  $\pi_{\theta} \leftarrow \pi_{\theta_0}$
  
  \tcp{Start the Loop}
  \For{$i=1$ \KwTo $T$}{
    \For{each $x$ in the input}{
    \tcp{\color{blue}{1-Online Exploration}}
  
    Generate $K$ symbolic solutions with self-rewards: $\{a_k\}_{k=1}^{K}, \{r_k\}_{k=1}^{K} \sim \pi_{\theta}(\cdot | x)$.
    
    Get binary rewards by executing in $\textbf{\textsc{Env}}$: $\{b_k\}_{k=1}^{K} \longleftarrow \mathbb{I}[{\textbf{\textsc{Env}}(a_k)==y}]$.
    
    Generate self-refined solutions with self-rewards: $\{\widetilde{a}_k\}_{k=1}^{K}, \{\widetilde{r}_k\}_{k=1}^{K} \sim \pi_{\theta}(\cdot | x;a_k)$.
    
    Get binary rewards by executing in $\textbf{\textsc{Env}}$: $\{\widetilde{b}_k\}_{k=1}^{K} \longleftarrow \mathbb{I}[{\textbf{\textsc{Env}}(\widetilde{a}_k)==y}]$.

    Let $T_k=(x, y, a_{k}, b_{k}, r_{k}), \widetilde{T}_k=(x, y, \widetilde{a}_{k}, \widetilde{b}_{k}, \widetilde{r}_{k})$ denote the collected trajectories.
    \vspace{0.3em}
    
    \tcp{\color{blue}{2-Traj. Filtering and Candidate Pool Updating}}

    Filter the superior trajectory $T_k^*$ from $T_k$ and $\widetilde{T}_k$ with binary rewards and self-rewards.
    
    Update the candidate pool with $T_k^*$.
    }
    
    \tcp{\color{blue}{3-Training}}

    Rank and retrieve positive-only training set $U_1$ and positive-negative pairs $U_2$ from $\textsc{POOL}$.
    \vspace{0.1em}
    
    Optimize $\pi_{\theta_0}$ to $\pi_{\theta}^*$ with $\mathcal{L} = -\sum\limits_{(x,a^+)\sim {U}_1}{\mathrm{log}} \, p_{\theta_0}(a^+|x) - \sum\limits_{(x,a^+,a^-)\sim {U}_2}{\mathrm{log}}\, p_{\theta_0}(a^+|x;a^-)$.

    Update the policy LLM for the next iteration: $\pi_{\theta} \leftarrow \pi_{\theta}^*$
  }
  \tcp{Output the enhanced LLM}
  Return  $\pi_{\theta}^*$\;
  \caption{A Neural-Symbolic Self-Training Framework \ours}
\end{algorithm}

% $T_k=(x, y, a_{k}, b_{k}, r_{k}), \widetilde{T}_k=(x, y, \widetilde{a}_{k}, \widetilde{b}_{k}, \widetilde{r}_{k})$

% \begin{table*}[t]
% \centering
% \footnotesize
% \resizebox{\linewidth}{!}{
% \begin{tabular}{l|ccccc|c}
%     \toprule
%     & Distill-GPT4 & Distill-Claude2 &Self-Reward &Iter. SFT+DPO &STaR+Env. & \textbf{\ours} \\
%     \midrule
%     \multicolumn{7}{c}{\cellcolor{gray!25} LLaMA2-Chat (7B)} \\
%     \midrule
%     Agent &\sim90h &\sim48h &\sim40h &\sim48h &\sim20h &\sim24h \\
%     Math Reason &\sim95h &\sim58h &\sim55h &\sim65h &\sim30h &\sim32h \\
%     Logic Reason &\sim28h &\sim18h &\sim35h &\sim40h &\sim20h &\sim24h \\
%     \midrule
%     \multicolumn{7}{c}{\cellcolor{gray!25} LLaMA2-Chat (13B)} \\
%     \midrule
%     Agent &\sim95h &\sim53h &\sim80h &\sim100h &\sim48h &\sim50h \\
%     Math Reason &\sim100h &\sim63h &\sim120h &\sim130h &\sim68h &\sim75h \\
%     Logic Reason &\sim32h &\sim23h &\sim80h &\sim95h &\sim45h &\sim50h \\
%     \midrule
%     \textbf{Total Cost} &18day (\textbf{+\$730}) &11day (\textbf{+\$585}) &17 day &20 day &10 day &11 day \\
%     \bottomrule
% \end{tabular}
% }
% \caption{Actual self-training time and financial cost.}
% \label{training_time}
% \end{table*}

\section{Definition of \emph{Exploratory Ability} and \emph{Stability}}
\label{appendix_definition}
(1) Whether the policy LLM can rapidly explore large amounts of correct samples, and (2) whether it can mitigate the issue of forgetting previously-solved samples are two key factors to evaluate LLMs in interacting with the environment. We define \emph{Exploratory Ability} (EA) and \emph{Stability} (STB) respectively as the metrics. The calculation of the metrics is defined as follows:

Suppose that we have the input set $M$. In the $i^{th}$ iteration, the solved sample (with correct trajectories) constitute of set $M_i$. 
$\bigcup_{j=1}^{i-1}{M_{j}}$ contains all the previously-solved samples from the iteration $1$ to $i-1$.
And $M_i \cup \bigcup_{j=1}^{i-1}{M_{j}}$ comprises the overlapped successful samples between the current iteration and preceding iterations.
$M_i \setminus \bigcup_{j=1}^{i-1}{M_{j}}$ denotes the sample set that are newly solved in the current iteration $i$.
Based on the definition, we accumulate to obtain the overall EA and STB of the entire process:
\begin{equation}
    \textrm{EA} = \sum\limits_{i=2}^{T} \frac{|M_i \setminus \bigcup_{j=1}^{i-1}{M_{j}}|}{|\bigcup_{j=1}^{i-1}{M_{j}}|}, \quad \textrm{STB} = \sum\limits_{i=2}^{T} \frac{|M_{i} \cap \bigcup_{j=1}^{i-1}{M_{j}}|}{|\bigcup_{j=1}^{i-1}{M_{j}}|}
\end{equation}

where $|\cdot|$ is the number of samples in the given set. $T$ is the total number of iterations.

Take the process of 2 iterations as an example, suppose the iteration 1 explores 1,000 correct samples. Iteration 2 obtains 1200 correct samples, including 800 previously-solved samples and 400 newly-solved samples. Then, $\textrm{EA}=400/1000$ and $\textrm{STB}=800/1000$.

\section{Supplementary Results}

\subsection{Evolution Progress}
\label{appendix_curves}
\begin{figure}[ht]
    \begin{subfigure}[b]{0.99\textwidth}
        \includegraphics[width=\textwidth, keepaspectratio]{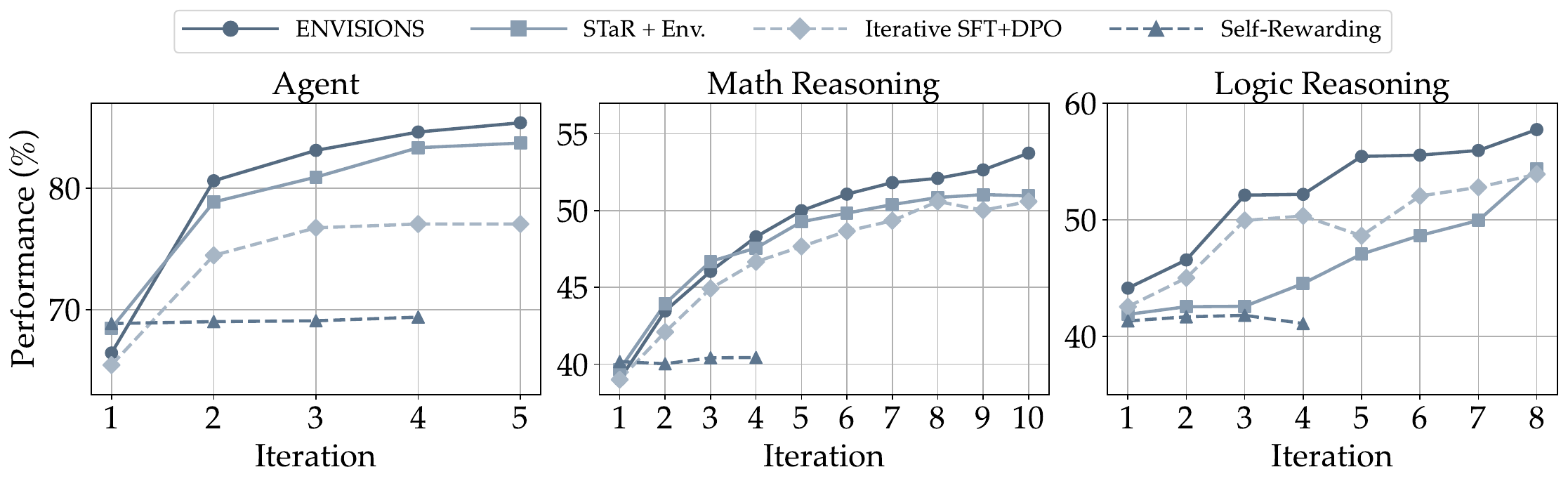}
        \caption{Evolution of performance.}
    \end{subfigure}

    \begin{subfigure}[b]{0.99\textwidth}
        \includegraphics[width=\linewidth]{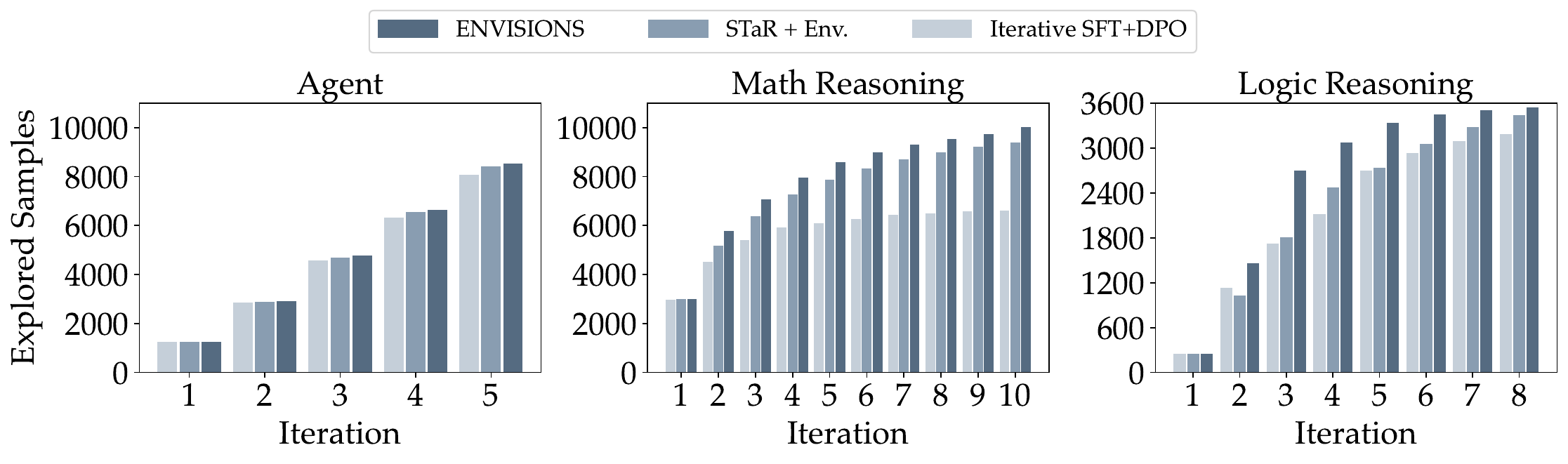}
        \caption{Evolution of explored sample numbers.}
    \end{subfigure}

    % \centering
    % \begin{subfigure}[b]{0.3\textwidth}
    %     \includegraphics[width=\textwidth, keepaspectratio]{Figures/sample_curve_llama2chat7b_agent.pdf}
    %     % \caption{Figure 1}
    % \end{subfigure}
    % \hfill
    % \begin{subfigure}[b]{0.34\textwidth}
    %     \includegraphics[width=\textwidth, keepaspectratio]{Figures/sample_curve_llama2chat7b_math.pdf}
    %     % \caption{Figure 2}
    % \end{subfigure}
    % \hfill
    % \begin{subfigure}[b]{0.34\textwidth}
    %     \includegraphics[width=\textwidth, keepaspectratio]{Figures/sample_curve_llama2chat7b_logic.pdf}
    %     % \caption{Figure 3}
    % \end{subfigure}
    % \vspace{-0.2cm}

    \caption{Evolution curves on LLaMA2-Chat 7B version across agent, math, and logic reasoning domains. (a) is the performance evolution progress. (b) denotes the evolution of explored sample numbers.}

    \label{appendix_exp_curves}
\end{figure}

Apart from the performance evolution curves with the LLaMA2-Chat 13B model presented in Figure~\ref{performance_curve}, we expand the discussion on the 7B version.
In Figure~\ref{appendix_exp_curves}, we visualize the evolution progress of self-training methods on both the model performance and the number of explored samples. The explored sample denotes that one input $x$ is solved by at least one generated symbolic solution $a_k$ (i.e., $b_k=1$). We count the number of explored samples at each iteration to make the figure.

% \begin{figure}
%     \centering
%     \includegraphics[width=\linewidth]{Figures/analysis/Merged_sample_curves_7B.pdf}
%     \caption{merged 7B}
%     \label{fig:merged_7b_samples}
% \end{figure}

From the results, the performances of the frameworks are positively correlated with the ability to continuously explore correct trajectories. \ours presents great superiority, especially in the logic reasoning tasks. Compared with our proposed \emph{Env-guided Self-Training} approach, \emph{Reinforced Self-Training} approach appears to be weaker at exploring new samples. This finding is consistent with Figure~\ref{analysis} in the main paper.

\subsection{Scaling of $K$}

\begin{table*}[h]
\centering
\footnotesize
\resizebox{\linewidth}{!}{
\begin{tabular}{l|c|ccccc|cc|c}
    \toprule
    \multirow{2}{*}{\textbf{Models}}& \multicolumn{1}{c|}{\textbf{Agent}} &\multicolumn{5}{c|}{\textbf{Math Reasoning}} &\multicolumn{2}{c|}{\textbf{Logical Reasoning}} &\multirow{2}{*}{\textbf{Avg.}} \\
    &\textbf{MiniWob++} &\textbf{GSM8K} & \textbf{MATH} & \textbf{GSM-H} &\textbf{SVAMP} &\textbf{ASDiv} &\textbf{ProofWriter} &\textbf{RuleTaker} & \\
    \midrule
    K=2 &78.56 &53.60 &17.37 &44.96 &67.20 &66.84 &35.17 &49.82 &51.69 \\
    \textbf{K=5} &\textbf{85.38} &\textbf{58.98} &19.00 &\textbf{48.52} &\textbf{72.40} &69.80 &52.83 &\textbf{62.63} &\textbf{58.69}  \\
    K=10 &79.24 &58.30 &21.89 &48.29 &67.90 &69.75 &53.50 &61.99 &57.61 \\
    K=15 &79.55 &57.47 &\textbf{23.72} &46.63 &69.80 &\textbf{70.28} &\textbf{54.83} &59.54 &57.73 \\
    \bottomrule
\end{tabular}
}
\caption{Scaling of $K$ with LLaMA2-Chat (7B) as the base LLM. In the main results, we implement $K=5$ for illustration.}
\label{appendix_exp_scale_k}
% \vspace{-1.25em}
\end{table*}

The hyper-parameter $K$ controls the number of generated candidate symbolic solutions at each generation step.
In the main results, we only implement $K=5$ for illustration.

In Table~\ref{appendix_exp_scale_k}, we present performances under various choices of $K$, including 2, 5, 10, and 15. Considering the training cost, we only include LLaMA2-Chat (7B) as the base LLM. From the results, we conclude the following takeaways:

\paragraph{Moderate value of $K$ leads to the optimal performances.}
When $K=5$, \ours demonstrates superior performances, especially on agentic tasks (i.e., MiniWob++ benchmark).
However, when reducing the value of $K$ (i.e., $K=2$), the overall performances of \ours drop a lot. It indicates that keeping a moderate number of candidate solutions in each generation step benefits the self-training process. 

\paragraph{Scaling of $K$ does not bring significant improvements.}
Scaling $K$ from 5 to 10 and 15 does bring improvements on some challenging tasks (e.g., MATH). However, this observation is not consistent across various tasks. Generally, the average performances remain stable with $K$ increasing.

% \subsection{Broader Advantages of \emph{Self-Refinement} in Inference-Time}

% \xfz{add some experiments on the inference-time self-refinement}

% \subsection{Analysis of Training Time}

% \xfz{refer to SimPO}

\subsection{MiniWob++ Results Per Tasks}
\label{appendix_miniwob}
Table~\ref{tab:miniwob} shows the performance of \ours on each of the 44 MiniWob++ tasks.

\begin{table*}[htbp]
\centering
\renewcommand\arraystretch{1.05}
\scriptsize
\scalebox{0.8}{%
\begin{tabular}{l|ccccccc}
\midrule
                            &\textbf{1-shot} & \textbf{Distill GPT4}    & \textbf{Distill Claude2}     & \textbf{STAR+Env.}    & \textbf{Self-Rewarding}   & \textbf{iter. SFT+DPO}     & \textbf{Ours}   \\ \hline
\multicolumn{8}{c}{\cellcolor{gray!25} LLaMA-2-Chat (7B)} \\
\midrule
choose-date                 & 0.00                      & 0.00                                              & 0.00                                                 & 0.00                                          & 0.00                                                 & 0.00                                             & 0.00                 \\
choose-list                 & 100.00                    & 100.00                                            & 100.00                                               & 100.00                                        & 100.00                                               & 100.00                                           & 100.00               \\
click-button                & 0.00                      & 100.00                                            & 100.00                                               & 100.00                                        & 100.00                                               & 96.67                                            & 100.00               \\
click-button-sequence       & 100.00                    & 100.00                                            & 100.00                                               & 100.00                                        & 100.00                                               & 100.00                                           & 96.67                \\
click-checkboxes            & 20.00                     & 100.00                                            & 96.67                                                & 100.00                                        & 100.00                                               & 100.00                                           & 100.00               \\
click-checkboxes-large      & 20.00                     & 86.67                                             & 96.67                                                & 86.67                                         & 66.67                                                & 100.00                                           & 100.00               \\
click-checkboxes-soft       & 0.00                      & 6.67                                              & 30.00                                                & 50.00                                         & 0.00                                                 & 63.33                                            & 76.67                \\
click-checkboxes-transfer   & 56.67                     & 100.00                                            & 100.00                                               & 100.00                                        & 100.00                                               & 100.00                                           & 100.00               \\
click-collapsible           & 100.00                    & 100.00                                            & 100.00                                               & 100.00                                        & 100.00                                               & 100.00                                           & 100.00               \\
click-color                 & 53.33                     & 100.00                                            & 100.00                                               & 100.00                                        & 100.00                                               & 100.00                                           & 100.00               \\
click-dialog                & 100.00                    & 100.00                                            & 100.00                                               & 100.00                                        & 0.00                                                 & 100.00                                           & 100.00               \\
click-dialog-2              & 0.00                      & 26.67                                             & 73.33                                                & 100.00                                        & 73.33                                                & 100.00                                           & 100.00               \\
click-link                  & 73.33                     & 93.33                                             & 93.33                                                & 93.33                                         & 93.33                                                & 93.33                                            & 93.33                \\
click-option                & 100.00                    & 100.00                                            & 100.00                                               & 100.00                                        & 100.00                                               & 100.00                                           & 100.00               \\
click-scroll-list           & 56.67                     & 100.00                                            & 100.00                                               & 100.00                                        & 96.67                                                & 100.00                                           & 100.00               \\
click-shades                & 93.33                     & 100.00                                            & 100.00                                               & 100.00                                        & 100.00                                               & 100.00                                           & 100.00               \\
click-shape                 & 0.00                      & 70.00                                             & 53.33                                                & 63.33                                         & 16.67                                                & 50.00                                            & 70.00                \\
click-tab                   & 100.00                    & 100.00                                            & 100.00                                               & 96.67                                         & 26.67                                                & 56.67                                            & 100.00               \\
click-test                  & 100.00                    & 100.00                                            & 100.00                                               & 100.00                                        & 100.00                                               & 100.00                                           & 100.00               \\
click-test-2                & 100.00                    & 100.00                                            & 100.00                                               & 100.00                                        & 100.00                                               & 100.00                                           & 100.00               \\
click-widget                & 96.67                     & 100.00                                            & 100.00                                               & 100.00                                        & 100.00                                               & 100.00                                           & 100.00               \\
copy-paste                  & 100.00                    & 100.00                                            & 100.00                                               & 100.00                                        & 100.00                                               & 100.00                                           & 100.00               \\
copy-paste-2                & 100.00                    & 100.00                                            & 100.00                                               & 100.00                                        & 100.00                                               & 100.00                                           & 100.00               \\
enter-date                  & 3.33                      & 100.00                                            & 100.00                                               & 100.00                                        & 100.00                                               & 100.00                                           & 100.00               \\
enter-password              & 96.67                     & 100.00                                            & 100.00                                               & 100.00                                        & 100.00                                               & 100.00                                           & 100.00               \\
enter-text                  & 100.00                    & 100.00                                            & 100.00                                               & 100.00                                        & 100.00                                               & 100.00                                           & 100.00               \\
enter-text-dynamic          & 100.00                    & 100.00                                            & 100.00                                               & 100.00                                        & 100.00                                               & 100.00                                           & 100.00               \\
enter-time                  & 0.00                      & 30.00                                             & 0.00                                                 & 0.00                                          & 0.00                                                 & 0.00                                             & 43.33                \\
focus-text                  & 100.00                    & 100.00                                            & 100.00                                               & 100.00                                        & 100.00                                               & 100.00                                           & 100.00               \\
focus-text-2                & 33.33                     & 100.00                                            & 100.00                                               & 100.00                                        & 100.00                                               & 100.00                                           & 100.00               \\
guess-number                & 6.67                      & 0.00                                              & 6.67                                                 & 10.00                                         & 6.67                                                 & 10.00                                            & 10.00                \\
identify-shape              & 0.00                      & 56.67                                             & 80.00                                                & 100.00                                        & 56.67                                                & 50.00                                            & 100.00               \\
multi-layouts               & 3.33                      & 96.67                                             & 86.67                                                & 100.00                                        & 76.67                                                & 96.67                                            & 100.00               \\
multi-orderings             & 0.00                      & 93.33                                             & 100.00                                               & 100.00                                        & 80.00                                                & 100.00                                           & 100.00               \\
navigate-tree               & 60.00                     & 60.00                                             & 60.00                                                & 60.00                                         & 60.00                                                & 60.00                                            & 60.00                \\
read-table                  & 70.00                     & 100.00                                            & 100.00                                               & 100.00                                        & 100.00                                               & 100.00                                           & 100.00               \\
search-engine               & 3.33                      & 100.00                                            & 100.00                                               & 100.00                                        & 43.33                                                & 0.00                                             & 100.00               \\
simple-algebra              & 6.67                      & 50.00                                             & 63.33                                                & 80.00                                         & 6.67                                                 & 3.33                                             & 73.33                \\
simple-arithmetic           & 0.00                      & 86.67                                             & 90.00                                                & 100.00                                        & 40.00                                                & 73.33                                            & 96.67                \\
social-media-all            & 30.00                     & 100.00                                            & 100.00                                               & 30.00                                         & 0.00                                                 & 30.00                                            & 30.00                \\
text-transform              & 66.67                     & 100.00                                            & 100.00                                               & 100.00                                        & 100.00                                               & 100.00                                           & 100.00               \\
unicode-test                & 100.00                    & 100.00                                            & 100.00                                               & 100.00                                        & 100.00                                               & 100.00                                           & 100.00               \\
use-slider                  & 0.00                      & 6.67                                              & 6.67                                                 & 6.67                                          & 6.67                                                 & 6.67                                             & 6.67                 \\
use-spinner                 & 0.00                      & 6.67                                              & 6.67                                                 & 6.67                                          & 6.67                                                 & 0.00                                             & 0.00                 \\
\midrule
\textbf{Average}           & 51.14     & 81.14                          & 82.80                                  & 83.71                         & 69.47                                & 77.05                            & \textbf{85.38} \\
\midrule
\multicolumn{8}{c}{\cellcolor{gray!25} LLaMA-2-Chat (13B)} \\
\midrule
choose-date                 & 0.00                      & 0.00                                              & 0.00                                                 & 0.00                                          & 0.00                                                 & 0.00                                             & 0.00                \\
choose-list                 & 96.67                     & 100.00                                            & 100.00                                               & 100.00                                        & 100.00                                               & 100.00                                           & 100.00              \\
click-button                & 96.67                     & 100.00                                            & 100.00                                               & 100.00                                        & 100.00                                               & 96.67                                            & 100.00              \\
click-button-sequence       & 100.00                    & 100.00                                            & 100.00                                               & 100.00                                        & 100.00                                               & 100.00                                           & 100.00              \\
click-checkboxes            & 30.00                     & 100.00                                            & 100.00                                               & 100.00                                        & 100.00                                               & 100.00                                           & 100.00              \\
click-checkboxes-large      & 26.67                     & 86.67                                             & 96.67                                                & 90.00                                         & 43.33                                                & 90.00                                            & 93.33               \\
click-checkboxes-soft       & 0.00                      & 3.33                                              & 46.67                                                & 90.00                                         & 20.00                                                & 60.00                                            & 90.00               \\
click-checkboxes-transfer   & 10.00                     & 100.00                                            & 96.67                                                & 100.00                                        & 100.00                                               & 100.00                                           & 100.00              \\
click-collapsible           & 100.00                    & 100.00                                            & 96.67                                                & 100.00                                        & 100.00                                               & 100.00                                           & 100.00              \\
click-color                 & 56.67                     & 100.00                                            & 100.00                                               & 100.00                                        & 100.00                                               & 100.00                                           & 100.00              \\
click-dialog                & 100.00                    & 100.00                                            & 100.00                                               & 100.00                                        & 0.00                                                 & 100.00                                           & 100.00              \\
click-dialog-2              & 0.00                      & 26.67                                             & 73.33                                                & 100.00                                        & 73.33                                                & 100.00                                           & 100.00              \\
click-link                  & 70.00                     & 93.33                                             & 93.33                                                & 93.33                                         & 93.33                                                & 93.33                                            & 93.33               \\
click-option                & 0.00                      & 100.00                                            & 100.00                                               & 100.00                                        & 100.00                                               & 100.00                                           & 100.00              \\
click-scroll-list           & 63.33                     & 100.00                                            & 100.00                                               & 100.00                                        & 100.00                                               & 100.00                                           & 100.00              \\
click-shades                & 100.00                    & 100.00                                            & 100.00                                               & 100.00                                        & 100.00                                               & 100.00                                           & 100.00              \\
click-shape                 & 10.00                     & 76.67                                             & 56.67                                                & 86.67                                         & 10.00                                                & 66.67                                            & 73.33               \\
click-tab                   & 100.00                    & 100.00                                            & 100.00                                               & 100.00                                        & 100.00                                               & 100.00                                           & 100.00              \\
click-test                  & 100.00                    & 100.00                                            & 96.67                                                & 100.00                                        & 100.00                                               & 100.00                                           & 100.00              \\
click-test-2                & 100.00                    & 100.00                                            & 100.00                                               & 100.00                                        & 100.00                                               & 100.00                                           & 100.00              \\
click-widget                & 96.67                     & 100.00                                            & 100.00                                               & 100.00                                        & 100.00                                               & 100.00                                           & 100.00              \\
copy-paste                  & 100.00                    & 100.00                                            & 100.00                                               & 100.00                                        & 100.00                                               & 100.00                                           & 100.00              \\
copy-paste-2                & 100.00                    & 100.00                                            & 100.00                                               & 100.00                                        & 100.00                                               & 100.00                                           & 96.67               \\
enter-date                  & 100.00                    & 100.00                                            & 100.00                                               & 100.00                                        & 100.00                                               & 100.00                                           & 100.00              \\
enter-password              & 100.00                    & 96.67                                             & 100.00                                               & 100.00                                        & 100.00                                               & 100.00                                           & 100.00              \\
enter-text                  & 100.00                    & 100.00                                            & 100.00                                               & 100.00                                        & 100.00                                               & 100.00                                           & 100.00              \\
enter-text-dynamic          & 100.00                    & 100.00                                            & 100.00                                               & 100.00                                        & 100.00                                               & 100.00                                           & 100.00              \\
enter-time                  & 0.00                      & 0.00                                              & 23.33                                                & 0.00                                          & 0.00                                                 & 0.00                                             & 93.33               \\
focus-text                  & 0.00                      & 100.00                                            & 100.00                                               & 100.00                                        & 96.67                                                & 100.00                                           & 100.00              \\
focus-text-2                & 63.33                     & 100.00                                            & 100.00                                               & 100.00                                        & 100.00                                               & 100.00                                           & 100.00              \\
guess-number                & 6.67                      & 0.00                                              & 6.67                                                 & 3.33                                          & 6.67                                                 & 10.00                                            & 6.67                \\
identify-shape              & 0.00                      & 10.00                                             & 90.00                                                & 100.00                                        & 20.00                                                & 90.00                                            & 100.00              \\
multi-layouts               & 66.67                     & 100.00                                            & 100.00                                               & 100.00                                        & 86.67                                                & 96.67                                            & 100.00              \\
multi-orderings             & 56.67                     & 100.00                                            & 100.00                                               & 100.00                                        & 100.00                                               & 100.00                                           & 100.00              \\
navigate-tree               & 60.00                     & 60.00                                             & 60.00                                                & 60.00                                         & 60.00                                                & 60.00                                            & 56.67               \\
read-table                  & 76.67                     & 100.00                                            & 100.00                                               & 100.00                                        & 100.00                                               & 100.00                                           & 100.00              \\
search-engine               & 90.00                     & 100.00                                            & 100.00                                               & 100.00                                        & 100.00                                               & 100.00                                           & 100.00              \\
simple-algebra              & 23.33                     & 76.67                                             & 80.00                                                & 80.00                                         & 43.33                                                & 36.67                                            & 83.33               \\
simple-arithmetic           & 56.67                     & 100.00                                            & 100.00                                               & 100.00                                        & 100.00                                               & 96.67                                            & 100.00              \\
social-media-all            & 93.33                     & 100.00                                            & 100.00                                               & 30.00                                         & 0.00                                                 & 30.00                                            & 30.00               \\
text-transform              & 96.67                     & 83.33                                             & 100.00                                               & 100.00                                        & 100.00                                               & 100.00                                           & 100.00              \\
unicode-test                & 93.33                     & 100.00                                            & 100.00                                               & 100.00                                        & 100.00                                               & 100.00                                           & 100.00              \\
use-slider                  & 0.00                      & 6.67                                              & 6.67                                                 & 6.67                                          & 6.67                                                 & 6.67                                             & 10.00               \\
use-spinner                 & 0.00                      & 6.67                                              & 6.67                                                 & 6.67                                          & 6.67                                                 & 6.67                                             & 6.67                \\
\midrule
\textbf{Average}                     & 60.00                        & 80.15                                       & 84.77                                          & 85.15                                   & 74.24                                                & 82.73                                            & \textbf{87.12} \\      
\midrule
\end{tabular}%
}
\caption{Detailed performances on 44 MiniWob++ tasks.}
\label{tab:miniwob}
\end{table*}

\end{document}